\documentclass[preprint,12pt]{elsarticle}
\usepackage{graphicx} 

\usepackage{algorithm}
\usepackage{algorithmicx}
\usepackage{algpseudocode}
\usepackage{amsmath}
\usepackage{amssymb}
\usepackage{mdwtab}
\usepackage{eqparbox}
\usepackage{makecell}
\usepackage{tabularray}
\usepackage[hidelinks]{hyperref}
\usepackage{lineno}
\usepackage{comment}
\usepackage{lscape}
\usepackage{multicol}
\usepackage{array}

\hypersetup{
    colorlinks=true,
    linkcolor=blue,
    filecolor=blue,      
    urlcolor=blue,
    citecolor=blue,
}

\usepackage[pdftex]{color}
\usepackage[font=footnotesize,labelfont=bf]{caption}
\usepackage[font=footnotesize,labelfont=bf]{subcaption}
\affiliation[label1]{organization={Concordia Institute for Information Systems Engineering, Concordia University},
            city={Montreal}, 
            country={Canada}}

\affiliation[label3]{organization={Department of Mechanical, Industrial and Aerospace Engineering, Concordia University},
            city={Montreal}, 
            country={Canada}}

\affiliation[label4]{organization={Department of Medicine, Laval University},
            city={Quebec}, 
            country={Canada}}

\affiliation[label5]{organization={Department of Computer Science, 6G Research Center, Khalifa University},
            city={Abu Dhabi}, 
            country={UAE}}

\affiliation[label6]{organization={Artificial Intelligence \& Cyber Systems Research Center, Department of CSM, Lebanese American University}, city={Beirut}, country={Lebanon}}

\author[label1]{Hanae Elmekki} 
\ead{hanae.elmekki@mail.concordia.ca}

 \author[label1]{Ahmed Alagha}
 \ead{ahmed.alagha@mail.concordia.ca}

\author[label1,label6]{Hani Sami}
\ead{hani.sami@mail.concordia.ca}

\author[label3]{Amanda Spilkin} 
\ead{amanda.spilkin@mail.concordia.ca}

\author[label4]{Antonela Mariel Zanuttini}
\ead{antonela-mariel.zanuttini.1@ulaval.ca}

\author[label3]{Ehsan Zakeri} 
\ead{ehsan.zakeri@concordia.ca}

\author[label5,label1]{Jamal Bentahar \corref{cor1}}
\cortext[cor1]{Corresponding Author's Email:  jamal.bentahar@concordia.ca}
\ead{jamal.bentahar@concordia.ca}

\author[label3]{Lyes Kadem}
\ead{lyes.kadem@concordia.ca}

\author[label3]{Wen-Fang Xie}
\ead{wenfang.xie@concordia.ca}

\author[label4]{Philippe Pibarot}
\ead{philippe.pibarot@med.ulaval.ca}

\author[label5]{Rabeb Mizouni}
\ead{rabeb.mizouni@ku.ac.ae}

\author[label5]{Hadi Otrok}
\ead{hadi.otrok@ku.ac.ae}

\author[label5]{Shakti Singh}
\ead{shakti.singh@ku.ac.ae}

\author[label5,label6]{Azzam Mourad}
\ead{azzam.mourad@lau.edu.lb}

\date{July 2024}

\begin{document}

\begin{frontmatter}

\title{CACTUS: An Open Dataset and Framework for Automated Cardiac Assessment and Classification of Ultrasound Images Using Deep Transfer Learning}

\begin{abstract}

Cardiac ultrasound (US) scanning is one of the most commonly used techniques in cardiology to diagnose the health of the heart and its proper functioning. During a typical US scan, medical professionals take several images of the heart to be classified based on the cardiac views they contain, with a focus on high-quality images. However, this task is time consuming and error prone. Therefore, it is necessary to consider ways to automate these tasks and assist medical professionals in classifying and assessing cardiac US images. Machine learning (ML) techniques are regarded as a prominent solution due to their success in the development of numerous applications aimed at enhancing the medical field, including addressing the shortage of echography technicians. However, the limited availability of medical data presents a significant barrier to the application of ML in the field of cardiology, particularly regarding US images of the heart. This paper addresses this challenge by introducing the first open graded dataset for Cardiac Assessment and ClassificaTion of UltraSound (CACTUS), which is available online. This dataset contains images obtained from scanning a CAE Blue Phantom and representing various heart views and different quality levels, exceeding the conventional cardiac views typically found in literature. Additionally, the paper introduces a Deep Learning (DL) framework consisting of two main components. The first component is responsible for classifying cardiac US images based on the heart view using a Convolutional Neural Network (CNN) architecture. The second component uses the concept of Transfer Learning (TL) to utilize knowledge from the first component and fine-tune it to create a model for grading and assessing cardiac images. The framework demonstrates high performance in both classification and grading, achieving up to 99.43\% accuracy and as low as 0.3067 error, respectively. To showcase its robustness, the framework is further fine-tuned using new images representing additional cardiac views and also compared to several other state-of-the-art architectures. The framework's outcomes and its performance in handling real-time scans were also assessed using a questionnaire answered by cardiac experts.
\end{abstract}


\end{frontmatter}

\section{Introduction}
\label{Sec:Introduction}
The medical field has greatly benefited from the rapid development of machine learning (ML) techniques, which have enhanced patient care effectively, enabled quicker healthcare services, and provided invaluable assistance to healthcare professionals in predicting diseases and making diagnoses \cite{javaid2022significance, may2021eight,kumari2023deep}. Among the specialties that have taken advantage of this integration is cardiology, which is a field focused on the diagnosis of the heart and related diseases \cite{averbuch2022applications,bhushan2023machine,petmezas2024recent}. One of the most widely used techniques for cardiac examination is ultrasound (US) imaging, referred to as echocardiography.
This cardiovascular imaging method, typically performed using transthoracic echocardiography (TTE), is vital for cardiac healthcare practitioners as it enables them to observe the heart's structure and function internally. It also provides insights into the shapes of heart chambers and the functionality of heart valves \cite{Sadeghpour2018}. Echocardiography helps in diagnosing several cardiac diseases related to the heart muscle and valves, such as heart failure, valvular heart diseases, and congenital heart defects, among others \cite{ciampi2007role,capotosto2018early}. 

The classification of US images according to the views of the heart is one of the common steps in the TTE examinations. However, categorizing cardiac views solely based on appearance presents challenges due to the variations in the heart structures, the US probe positioning, and the differences in the inherent heart characteristics \cite{balaji2015automatic}. Additionally, the process of acquiring high-fidelity cardiac US images is predominantly reliant upon the expertise of skilled medical practitioners, which need to meticulously manipulate the US probe to obtain comprehensive views of the heart from different angles/views, and this can be challenging for those with limited expertise in the field \cite{pop2020classification}.

To address the aforementioned challenges, the recent advancements in Deep Learning (DL) and Artificial Intelligence (AI) provide varying solutions to assist medical practitioners in acquiring high quality and informative cardiac images with less time and fewer errors. DL can be used effectively to classify and evaluate the cardiac images, providing real-time feedback to users. In this regard, the literature on cardiac US encompasses numerous applications of DL techniques aimed at supporting medical professionals and providing AI-guided assistance in their field. Several studies \cite{ostvik2019real,blaivas2020all,gao2021automated} propose the use of Convolutional Neural Networks (CNNs) to classify the cardiac views and aid the medical practitioners in capturing the desired views. These works consider and compare different CNN architectures, as well as different variations of the cardiac views. This classification process is crucial for ensuring continuous scanning and real-time analysis of multiple quantitative parameters during TTE process without relying solely on highly experienced echocardiographers. The existing proposals in the literature have the common challenge of the lack of readily available public US image datasets, particularly those focused on cardiac imaging, to train and test DL models, which explains often the utilization of private datasets for research purposes. This constraint in data availability can be linked to various factors, such as the sensitive nature of medical data and the complexity involved in processing medical images \cite{deheyab2022overview}. Additionally, the aforementioned proposals focus solely on the classification of US cardiac views, with no attention given to the quality assessment of the image within each view. Generally, US image clarity is frequently compromised by speckle-induced blurring, impeding accurate interpretation and quantitative analysis, thereby affecting clinical diagnosis accuracy \cite{bharti2020ultrasound}. Furthermore, variations in probe positioning and orientation can influence the US image completeness and depiction of targeted cardiac structures. Hence, achieving optimal image quality across different probe positions is essential for enhanced diagnosis, an aspect often overlooked in existing proposals. While various approaches, including DL models and expert evaluations \cite{zhang2021cnn,wu2017fuiqa,adamson2020development}, have been explored to evaluate medical US images, there is currently no publicly available graded dataset of cardiac US images that can be generally utilized to develop new DL models.
As a result, this advocates for a comprehensive and publicly available dataset that also considers the quality (or grading) of US images within each cardiac view that reflects their completeness and clarity from noise, which is missing from the literature. In summary, the drawbacks of existing works addressing the classification of US cardiac images are: 
\begin{itemize}
\item There are limitations in the availability of public datasets of US cardiac scans to train DL models.
\item Lack of data representing graded cardiac scans that consider the quality and informativeness of the collected images, particularly in terms of cardiac structure completeness and clarity.
\item Existing AI-based US cardiac image classification proposals do not consider the quality of the image when guiding the medical practitioner.
\end{itemize}

In this paper, we aim to tackle the aforementioned challenges in the existing literature by introducing the dataset for Cardiac Assessment and ClassificaTion of Ultrasound (CACTUS). This dataset is the first of its kind, offering a comprehensive collection of graded cardiac views assessed by cardiovascular imaging experts and made publicly available. As a first step towards this, the images are collected by scanning the CAE Blue Phantom\footnote{\url{https://medicalskillstrainers.cae.com/cardiac-ultrasound-training-block/p?skuId=30}} replicating the human heart. A detailed grading schema is created and used to assess and grade each image through medical professionals. Using the CACTUS dataset, an AI-based framework is developed for classifying and grading US cardiac images, leveraging advanced DL techniques for improved accuracy and performance.

\begin{figure}[h]
  \begin{subfigure}{.30\textwidth}
  \centering
    \includegraphics[width=1\linewidth]{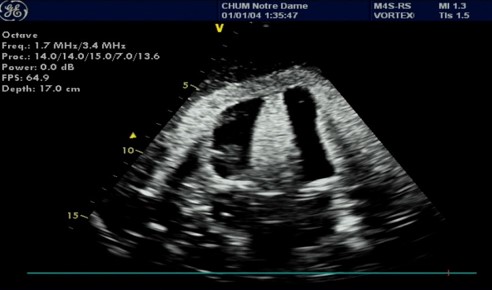}
    \caption{Uncompleted Cardiac View}
  \end{subfigure}%
  \hspace{1em} %
  \begin{subfigure}{.30\textwidth}
  \centering
    \includegraphics[width=1\linewidth]{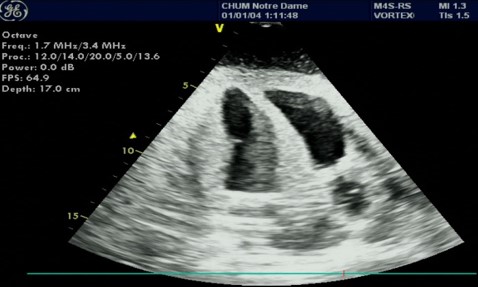}
    \caption{Speckled Cardiac View}
  \end{subfigure}%
  \hspace{1em} %
  \begin{subfigure}{.30\textwidth}
  \centering
    \includegraphics[width=1\linewidth]{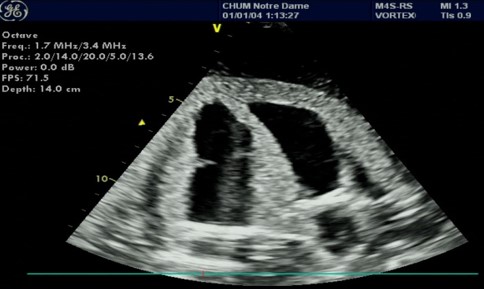}
    \caption{Completed Cardiac View}
  \end{subfigure}
  \caption{Different images of the apical four chamber cardiac view}
  \label{fig:a4c}
  \vspace{-1em}
\end{figure}

However, the development and deployment of DL models in such applications faces many challenges. To begin with, recognizing cardiac views in the TTE process is challenging due to subtle differences among various views \cite{madani2018fast}. Figure \ref{fig:a4c} illustrates three images of the apical four chamber view, a standard view of the heart. These images depict the same view but exhibit variances stemming from noise, speckles, and differences in US scanning machine and probe configurations. 
Furthermore, the DL models is a computationally expensive task. It is essential to ensure quick performance with low memory and computational complexity for efficient deployment. Additionally, due to the dynamic nature of the cardiac US scanning and the different variations with different patients, it is important to ensure the possibility of improving the developed DL models over time with more data. In summary, the following are the challenges we address when building our proposed framework:

\begin{itemize}
    \item The subtle difference between the different cardiac views makes it challenging for the DL model to differentiate between them.
    \item There is an increased computation complexity if we build two separate models for classification and grading.
    \item The CNN model is supposed to be adaptable to generalize for new patients' heart views.
\end{itemize}

Since both the classification and grading models are expected to handle images with similar features, and to mitigate computational complexities associated with the use of multiple DL models, we utilize Transfer Learning (TL) to build a common feature extractor with two heads: one for classification and one for grading. The classification head is responsible for identifying the cardiac view, while the grading head is responsible for measuring the informativeness and the usability of the cardiac image within that view. Using this approach, the model is further fine-tuned using new cardiac views added to the dataset to explore its adaptability. The main contributions of this study can be summarized as follows:

\begin{itemize}
        \item Introducing the first publicly available dataset comprising graded cardiac US images, representing diverse heart views.
        \item Developing a training framework combining classification and grading models using TL-assisted DL architecture, while also studying the adaptability of the models through fine-tuning on new data.    
        \item Conducting a comparative analysis of the proposed framework with existing cutting-edge CNN architectures, while testing the TL capabilities.
        \item Perform real-time scans using the proposed framework and assess its results through feedback from cardiac imaging experts. The framework's performance is further evaluated with a questionnaire completed by two cardiac imaging expert.
\end{itemize}

The remainder of this paper is structured as follows: Section \ref{Sec: RelatedWork} provides an extensive review of existing literature concerning cardiac US datasets and the utilization of DL with cardiac US images. Section \ref{subsec: CACTUS dataset} presents the proposed dataset, encompasses various graded views of the heart. Section \ref{Sec: Proposed Framework} details the TL-assisted DL framework for classifying and grading cardiac views. Section \ref{sec: Experiments} focuses on the experimental setup and simulation results to evaluate the proposed methods. Finally, Section \ref{sec: concolusions} concludes the paper by summarizing our findings and discussing potential directions for future research.

\section{Related Work}
\label{Sec: RelatedWork}
The purpose of this section is to survey and analyze existing literature concerning public datasets and DL classification for cardiac US images.

\subsection{Public Cardiac US Datasets}
\label{subsec: RW datasets}

Publishing datasets in the literature is crucial for facilitating the development of robust and innovative DL models, as well as for enabling the comparison of different models using a common benchmark. There are numerous publicly available US datasets for various human organs in the literature. One example is the COVIDx-US dataset \cite{DBLP:journals/corr/abs-2103-10003}, which contains US imaging data of lungs related to COVID-19 infections. Another instance is presented in \cite{pedraza2015open}, which introduces a dataset of thyroid US images, annotated and described with a focus on lesions by medical experts. Additionally, \cite{DBLP:conf/miccai/SinglaRHLRNR23} has introduced the first publicly available kidney US dataset, collected from over 500 adult patients and annotated by sonography experts for DL applications. Similarly, \cite{gomez2023bus} presents BUS-BRA, a publicly accessible breast US dataset designed for lesion detection, segmentation methods, and classification of breast US images.

However, the availability of public US cardiac images is limited. Through our literature review, we identified two primary datasets: CAMUS \cite{DBLP:journals/tmi/LeclercSPOCEEBJ19} and EchoNet-Dynamic \cite{DBLP:journals/nature/OuyangHGYELHHLA20}. The CAMUS Dataset consists of 2000 labeled US images displaying apical four chamber and two chamber views, sourced from 500 patients and meticulously labeled by three cardiologists. This dataset can be accessed at the CAMUS Dataset website. On the other hand, the EchoNet-Dynamic dataset comprises 10,030 labeled apical four chamber echocardiogram videos, each accompanied by labels from human experts. This dataset is accessible through the EchoNet-Dynamic website. Although existing datasets offer cardiac data, they predominantly feature limited variations of cardiac views, particularly focusing only on the apical four chamber and apical two chamber views. Moreover, these images lack grading or assessment for quality, rendering them unsuitable for training DL models to evaluate the quality of such images.

In fact, there is a lack of standardized image quality assessment (IQA) due to its subjective nature \cite{zhang2021cnn,hemmsen2010ultrasound}. Among the studies that have been proposed in order to address this issue, the authors in \cite{adamson2020development} offer a checklist of criteria that should be considered in order to evaluate the acquisition of focus cardiac US (FoCUS) images. Despite the absence of standardized US image assessment criteria, many studies have investigated the evaluation of medical US images through the utilization of DL models \cite{zhang2021cnn,wu2017fuiqa}, as well as through established evaluation criteria by medical professionals. Nevertheless, there is currently no publicly available graded collection of cardiac US images that can be generally utilized to develop new DL models.

Our proposed dataset offers a wide array of views beyond the typical ones, including subcostal, apical four chamber, parasternal long axes, and parasternal short axes, which are crucial for accurate cardiac diagnosis. The collected images are also graded according to a scale created by cardiovascular imaging experts who meticulously assess US images based on two criteria: clarity and completeness. Furthermore, the CACTUS dataset is continually expanding and can accommodate a vast number of images since it is generated by scanning a phantom that simulates a heart, eliminating the need for specific configurations and ethical considerations.

\subsection{Deep Learning for Cardiac US}
\label{subsec: RW DL for US}
Despite encountering challenges associated with speckles, noise, low contrast, and variability in cardiac structures present in US images, there are several proposals in the literature that explore the application of DL on cardiac US images. The authors in \cite{DBLP:journals/bspc/SfakianakisST23,DBLP:journals/mia/LiuWLYT21} use DL to segment the cardiac US images and apply quantitative measurement, using the CAMUS \cite{DBLP:journals/tmi/LeclercSPOCEEBJ19} and EchoNet-Dynamic \cite{DBLP:journals/nature/OuyangHGYELHHLA20} datasets, respectively. Some of the works attempt to overcome the data limitations by using data augmentation techniques to extrapolate the data into new ones. In a similar context, the authors in \cite{DBLP:journals/cmig/XuLSWLWWLYHGZH20} created a private dataset of US images to perform semantic segmentation of cardiac views. They employed a CNN for US segmentation to identify the anatomical structures of the apical four-chamber view, aiming for early detection of congenital heart disease (CHD). In order to estimate the left atrial volume (LAV) for predicting specific undesirable cardiovascular events, the authors in \cite{barzegar2021proposing} employ a CNN model for classification and grading tasks to identify the cardiac cycle phase and estimate the maximal and minimal LAV without the need for image segmentation. The study was based on the collection of images of apical four chamber views from 2D echocardiography videos and incorporated data augmentation techniques. To further enhance DL applications in cardiac US imaging, some works have turned to simulated data to address challenges associated with the lack of medical data. For instance, certain works have proposed simulating color Doppler data to support DL models in Doppler velocity estimation \cite{puig2024boosting}, while others have developed clinical-like US cardiac sequences for training DL models in cardiac reconstruction and motion estimation \cite{lu2023ultrafast}. Other proposals focus on using DL to classify cardiac views, with the aim of aiding medical professionals in diagnosing cardiac diseases. The authors in \cite{ostvik2019real} conducted a comparison of three CNNs using a private dataset constructed by scanning both unhealthy and healthy patients, with the goal of classifying seven cardiac views within US images and achieving notable levels of accuracy. Similarly, in the realm of computer-aided recognition, another study \cite{gao2021automated} utilized a CNN model to automatically classify nine cardiac views issued from two private datasets collected by scanning 700 patients. Moreover, with a focus on Point-of-Care US (POCUS) as a solution for rapid decision-making in urgent care scenarios, \cite{blaivas2020all} evaluated the effectiveness of six distinct CNN architectures trained on images depicting five cardiac views, including parasternal long axis, parasternal short axis, apical four chamber, subcostal four chamber, and subcostal inferior vena cava, extracted from publicly available open-access US videos. The analysis from this research indicates that shallower models tend to outperform deeper ones in terms of performance. Different from previously cited studies, this study \cite{zamzmi2022real} proposes a solution to classify US images and assess their quality (in addition to image segmentation and quantification) using a MobileNetV2-based autoencoder. For training and fine-tuning the model, they used four datasets, including EchoNet-Dynamic and three more private datasets. The classification includes four classes: apical four chamber, inferior vena cava, parasternal long axis, and Doppler ultrasounds. Moreover, the evaluation of image quality uses a binary classifier that categorizes images as either bad or good quality.

The proposed framework addresses the lack of available data by introducing the first publicly available and graded dataset of cardiac US images. Additionally, the DL component of the framework is able to classify images from the CACTUS dataset, which includes representations of five cardiac views of varying degrees of quality alongside random images, to enhance the performance of the trained model. Additionally, our framework differs from previous proposals by its capability to evaluate images through the application of TL, which is a significant approach in enhancing the learning process via transfer knowledge. In our case, TL allows us to reuse parts of the previously trained classification model to grade images, thus saving prediction time and computational complexity. To demonstrate the robustness of the proposed framework, this latter is fine-tuned once again using the TL approach on a new cardiac view that was not introduced during training, yielding promising results.

\section{The CACTUS Dataset}
\label{subsec: CACTUS dataset}

\begin{figure}[h]
    \begin{center}
    \centering
\includegraphics[width=0.7\linewidth]{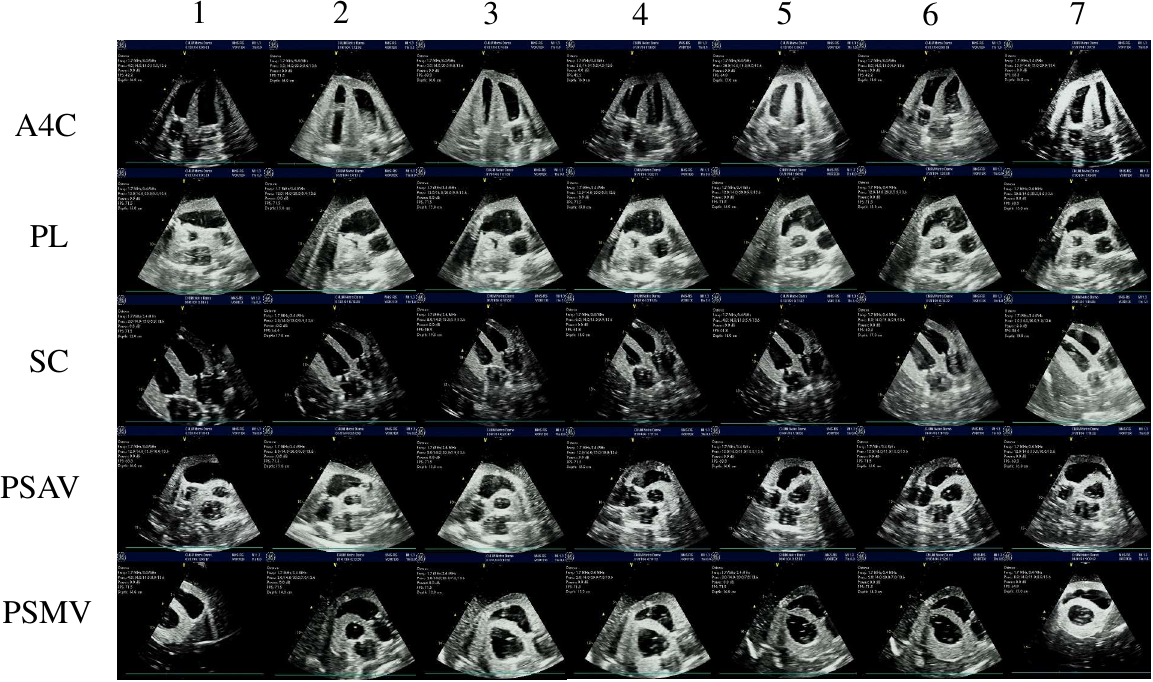}
    \caption{Graded cardiac views}
    \label{fig:cardiac_views}
    \end{center}
    \vspace{-2em}
\end{figure}

The human heart, with its intricate structure and shape, presents a complex organ. Tools like the CAE Blue Phantom offer a solution to replicate the human heart until the acquisition of real human heart scans. Utilizing cardiac phantom scans provides an efficient means to collect comprehensive data on various heart views. Considering this aspect, we establish a dataset that comprises five cardiac views, which are fundamental views of the heart: apical four chamber (A4C), subcostal four chamber (SC), parasternal long axis (PL), and two parasternal short axis views - one for the aortic valve (PSAV) and the other for the mitral valve (PSMV). Moreover, the dataset contains another class of data comprising a selection of random US images, depicting arbitrary probe positions that do not correspond to specific cardiac views. These images are beneficial for training DL models as they enable the model to differentiate between cardiac views and random images when fed a stream of images from an ongoing scan. Each image in the dataset is assigned a grade by cardiac imaging experts on a scale of 1 (poor quality) to 10 (excellent quality). The random images are assigned a grade of 0 since they do not represent any specific cardiac view. Figure \ref{fig:cardiac_views} illustrates some sample views encompassed within the dataset graded from 1 to 7. It can be seen that for the same view, the higher the grade the more visible the features representing the view are. Consequently, it is worth mentioning that sometimes a very minor probe movement could significantly affect the image quality, highlighting the complexity of the process.

\subsection{Data Acquisition} 
\label{subsubsec: Data Acquisition}

\begin{figure}[h]
    \begin{center}
    \centering
\includegraphics[width=0.6\linewidth]{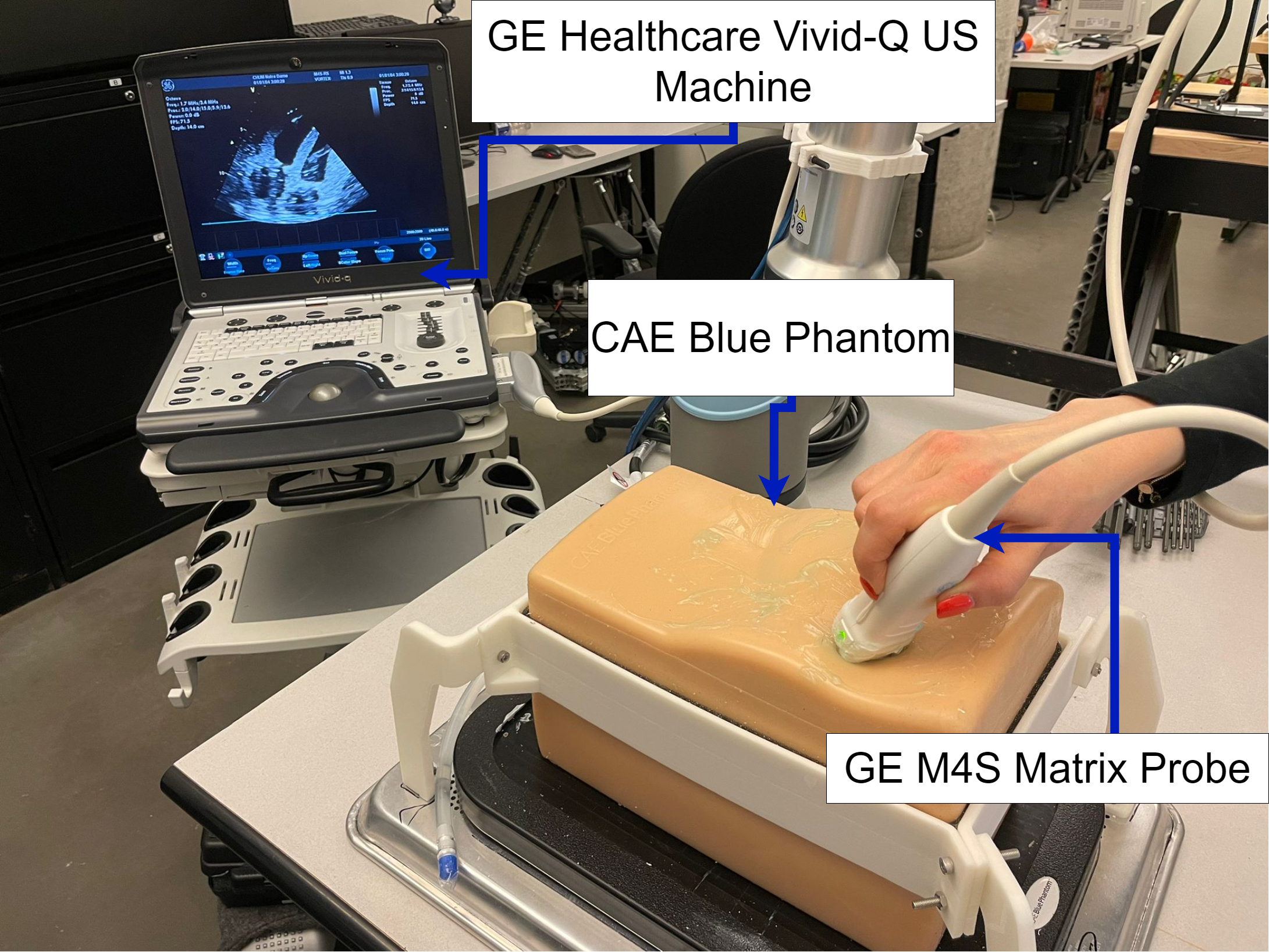}
    \caption{An illustration of the CAE Blue Phantom scanning setup}
    \label{fig:setup}
    \end{center}
    \vspace{-1em}
\end{figure}

\begin{figure}[H]
    \begin{center}
    \centering
\includegraphics[width=0.9\linewidth]{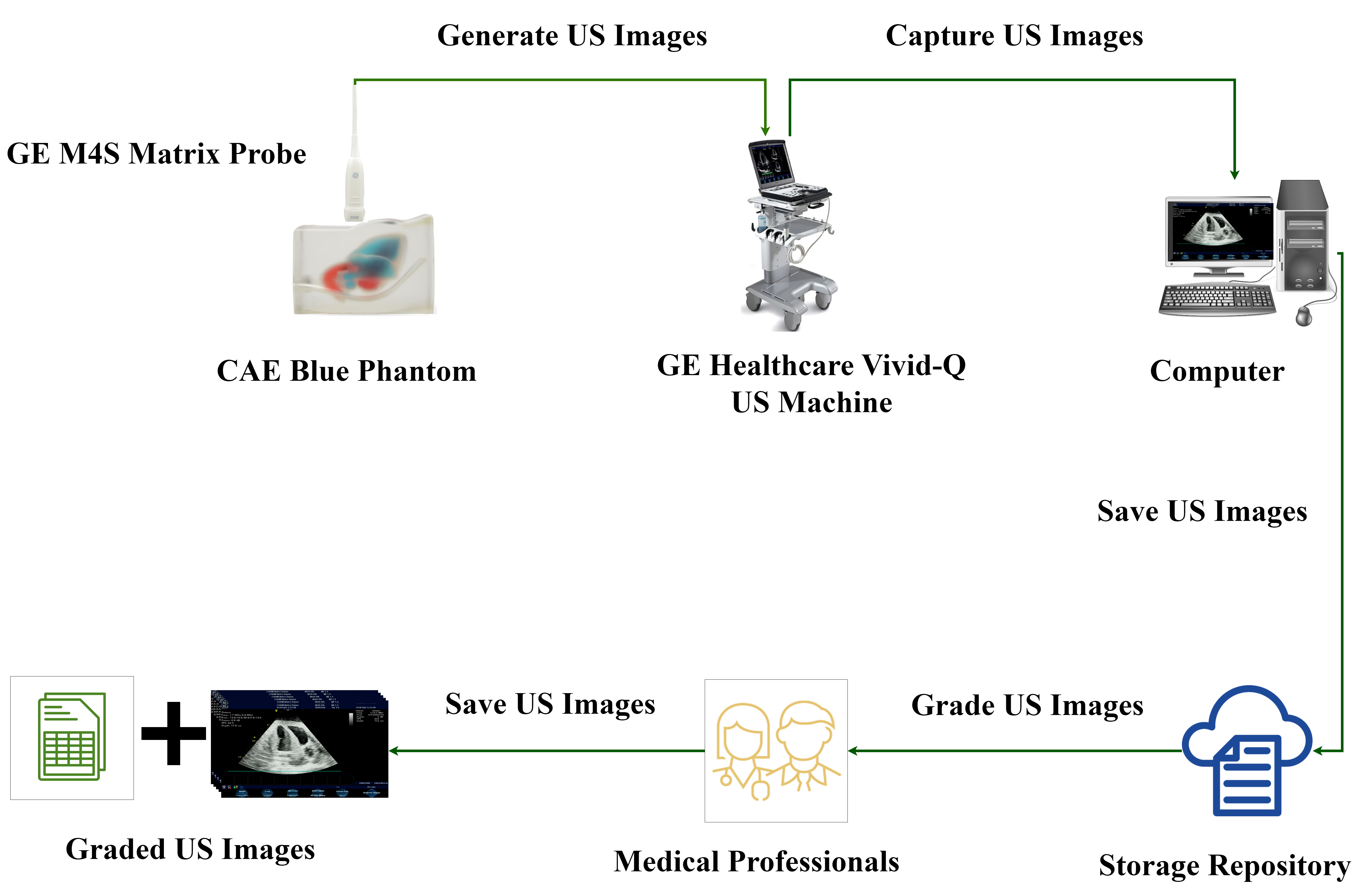}
    \caption{Data acquisition process}
    \label{fig:acqui}
    \end{center}
    \vspace{-1em}
\end{figure}

In this study, we conducted the scanning process on the CAE Blue Phantom using the GE M4S Matrix Probe\footnote{\url{https://services.gehealthcare.com/gehcstorefront/p/5499597}} and the GE Healthcare Vivid-Q US machine\footnote{\url{https://services.gehealthcare.com/gehcstorefront/}}. The phantom scanning process, as depicted in Figure \ref{fig:setup}, produces US images that are saved in a computer linked to the US machine. The images are categorized into the predefined cardiac views and are then stored in a repository for evaluation by skilled cardiovascular imaging experts. Subsequently, our proposed framework is used to train DL models on the CACTUS dataset to classify and grade the different views, offering vital support to medical practitioners in their cardiac diagnostic and examination endeavors. A detailed illustration of this image acquisition procedure is provided in Figure \ref{fig:acqui}.

\begin{figure}[H]
    \begin{center}
    \centering
\includegraphics[width=.6\linewidth]{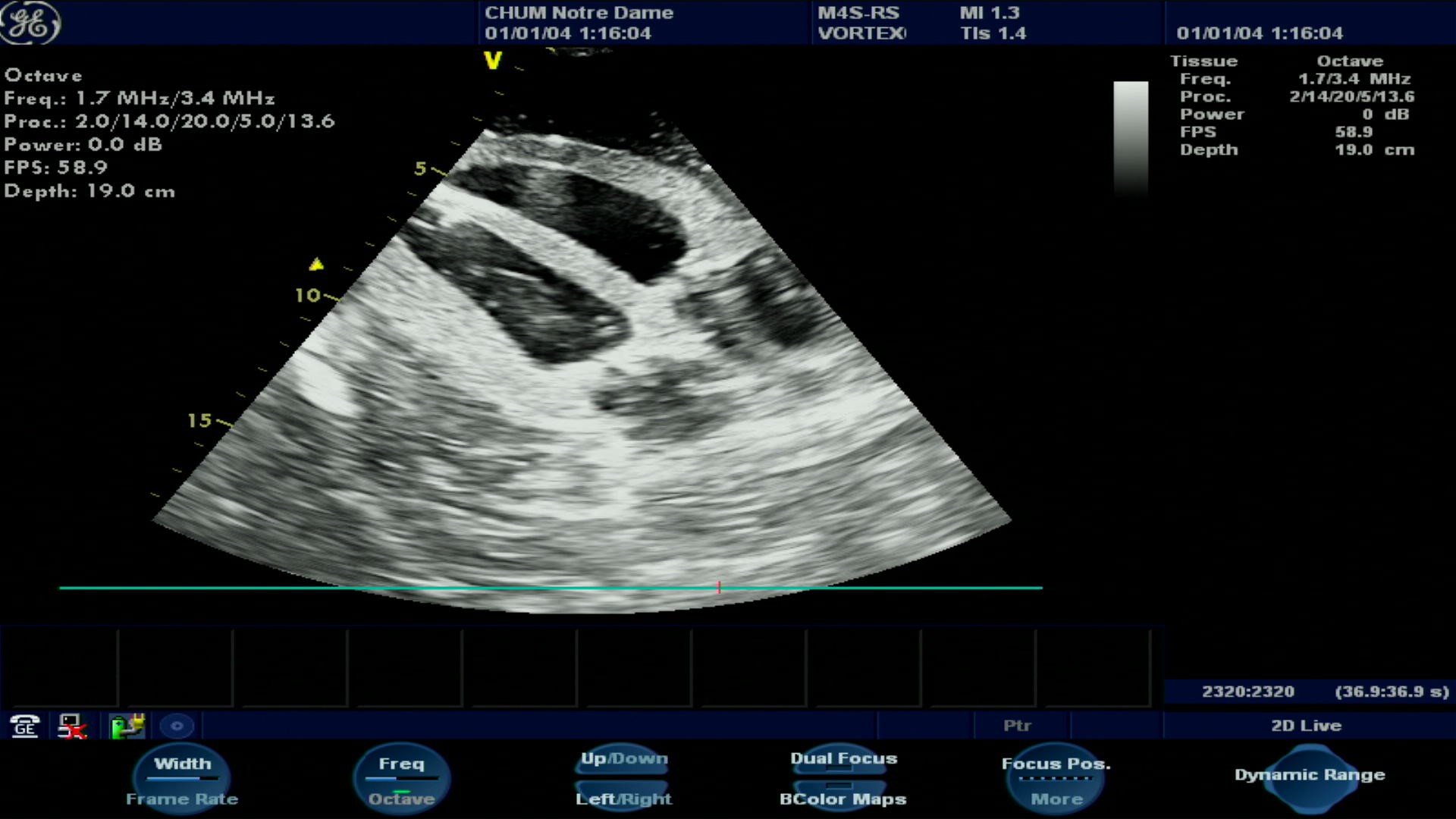}
   \caption{Subcostal four chamber cardiac view}
    \label{fig:sc_view}
    \end{center}
    \vspace{-1em}
\end{figure}

During US scanning, a multitude of parameters must be considered to achieve optimal scanning results. Throughout the scan of the CAE Blue Phantom, we carefully selected multiple parameters to achieve optimal and clear views of the targeted structures. These parameters, including depth, gain, dynamic range, frequency and power, are fundamental for guiding a cardiac US examination. We documented the range of values for these parameters in Table \ref{tab:us_parameters_tuning}, carefully varying them to produce images ranging from poor to high quality. Indeed, modifying these parameters influences US image quality, and the adjustments are contingent upon the particular application for which the images are intended. For instance, reducing the gain parameter may lead to the presence of speckles and noise within the image. From another side, elevating gain enhances image brightness, whereas augmenting dynamic range diminishes image brightness, thus indicating that dynamic range and gain settings jointly impact image brightness. Additionally, modifications in depth and frequency can impact image resolution; raising frequency typically enhances resolution, while depth values are proportionally related to image resolution. As an example, Figure \ref{fig:sc_view} depicts an US image obtained from scanning the phantom, showing the subcostal four chamber view. As illustrated in this figure, the image is captured under the following configuration: depth set to 19 cm, gain adjusted to 2 db, frequency set at 58.9 Hz, probe frequency at 1.7 MHz, dynamic range set at 5 db and power set to 20 W. It is worth mentioning that, for the generated dataset, different variations and combinations of these parameters have been used to produce the images.

\begin{table}[H]
  \centering
  \begin{tabular}{|c |c|}
\hline
    US parameter & Value Range \\ \hline

Depth & 13 - 17 cm \\
\hline

Gain & (-30) to 30 db \\
\hline

Dynamic Range & 1 - 10 db  \\
\hline

Power & 2 - 20 W  \\
\hline

Machine Frequency &  71.5 Hz \\
\hline

Probe Frequency &  1.5 - 3.6 MHz \\
\hline

Data Acquisition Rate &  30 Frames per second (fps) \\
\hline

\end{tabular}
  \caption{US parameters tuning}
\label{tab:us_parameters_tuning}
\end{table}

\subsection{Grading Process}
\label{subsubsec: Grading Process}

Evaluating cardiac US images is crucial for analyzing images obtained from the CAE Blue Phantom and constructing a dataset featuring diverse levels of image quality. Within the scope of this study, the CACTUS dataset is evaluated by imaging experts, who have created a grading system centered on two key factors: completeness and clarity. Completeness evaluates the visibility of the targeted cardiac structures in the image, assigning higher grades to images displaying the entire structure compared to those revealing only partial views. Clarity examines the luminosity of images and their purity from speckles and noise. The grading scale spans from 0 to 10, where 0 signifies an image that fails to capture a specific cardiac window, rendering it uninterpretable, whereas 10 represents a fully visible cardiac view with distinctly identifiable structures and optimal gain/power settings for clear delineation. This single-score approach, which combines multiple criteria based on a predefined scale, is commonly used in the literature \cite{abdi2017automatic,gaspari2021development}, where image quality rating scales ranging from 1 to 5 or 0 to 5 are applied to evaluate US images from TTE scans. Table \ref{tab:garding_schema} provides additional insights into the grading methodology, and the image sample corresponds to the one previously displayed in Figure \ref{fig:cardiac_views}, which illustrates the images for the various classes along with their associated grades.

\begin{table}
  \centering
  \begin{tabular}{|p{1.5cm} |p{11cm}|}
\hline
    Grade (G) & Description \\ \hline
0$\leq$G$<$1 &
The image does not capture a specific cardiac window and does not permit for interpretation of cardiac structures\\
\hline

 1$\leq$G$<$2 & The cardiac view is lightly visible, with one or more cardiac structures missing or obstructed, possibly due to improper depth settings, blurry structures, undefined borders, and high noise levels, indicative of insufficient gain/power.
 \\
\hline

2$\leq$G$<$3 &
The cardiac view in the image is partially visible, with some cardiac structures being only partially interpretable or missing, suggesting inadequate gain/power for clear delineation of cardiac structures. Presence of significant obstructions in the view.
\\
\hline

3$\leq$G$<$4 & A partial view of the cardiac anatomy can be seen, with some cardiac structures only partially discernible or not visible, indicating inadequate gain/power for clear delineation of cardiac structures. Presence of fewer obstructions.

\\
\hline

 4$\leq$G$<$5 & The majority of the cardiac view is visible and  cardiac structures are visible but may be obstructed. The gain/power is satisfactory to obtain a clear delineation of cardiac structures.
\\
\hline

 5$\leq$G$<$6 & The cardiac image presents a clear view and identification of cardiac structures, with a slight chance of minor obstructions, indicating adequate gain/power for clear delineation.
\\
\hline

 6$\leq$G$<$7 & The cardiac structures are visible and identifiable, with minimal chance of minor obstructions, indicating sufficient gain/power for clear delineation.
\\
\hline

 7$\leq$G$\leq$10 & 
The cardiac view is fully visible, with structures in the image clearly identifiable. The grading primarily relies on parameter tuning: grade 7 indicates that the gain/power is nearly optimal for clear delineation of cardiac structures, grade 8 is more optimal, grade 9 indicates even better gain/power, and grade 10 signifies optimal gain/power.
\\
\hline

\end{tabular}
  \caption{Grading schema}
\label{tab:garding_schema}
\end{table}

\subsection{Dataset Statistics}
\label{subsubsec: Dataset Stats}

\begin{figure}[h]
    \begin{center}
    \centering
\includegraphics[width=0.8\linewidth]{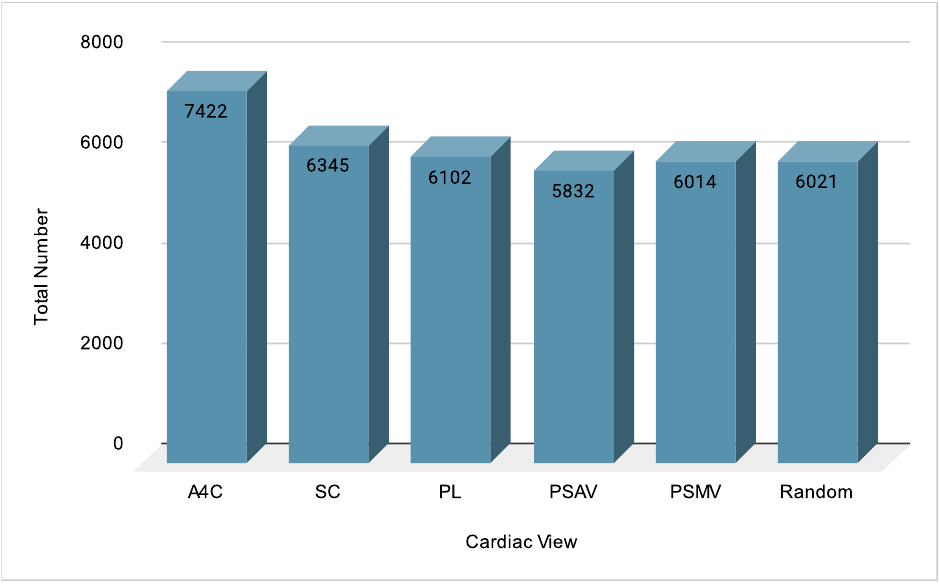}
   \caption{Dataset statistics}
    \label{fig:dataset_statistics}
    \end{center}
    \vspace{-2em}
\end{figure}

The CACTUS dataset comprises a total of 37,736 cardiac US images classified and graded on a scale from 0 to 10, as illustrated in Figure \ref{fig:dataset_statistics} which depicts the distribution of views within the dataset.
The dataset contains five cardiac views, namely A4C, SC, PL, PSAV, and PSMV. The sixth class is a set of random US images collected under varying probe positions. The dataset can be accessed via this link: \href{https://users.encs.concordia.ca/~kadem/cactus/}{\underline{CACTUS Dataset}}.

\section{Proposed Framework}
\label{Sec: Proposed Framework}
\begin{figure}[h]
    \begin{center}
    \centering
\includegraphics[width=1.0\linewidth]
{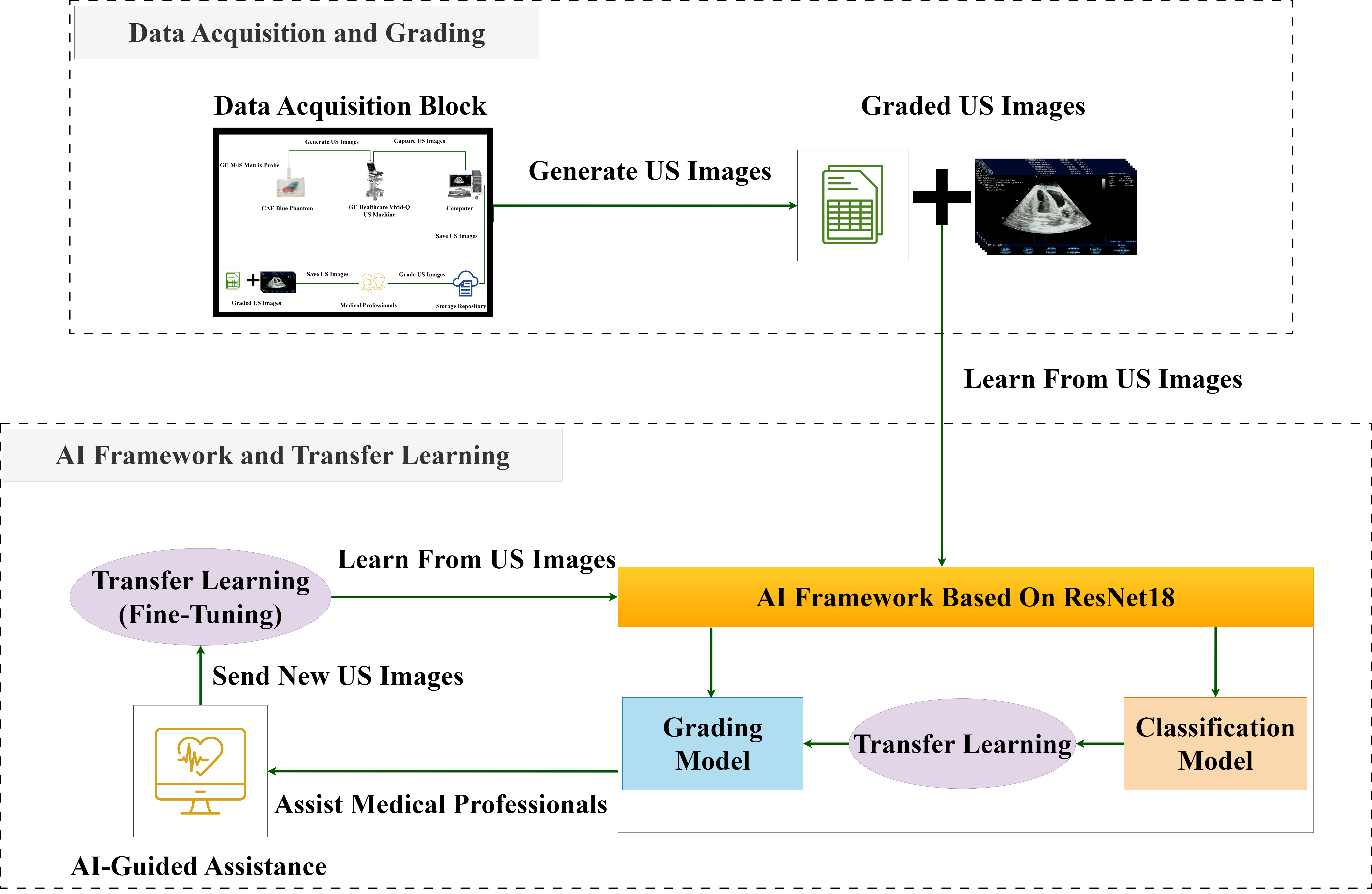}
    \caption{Overall process of the proposed solution (Section \ref{subsubsec: Data Acquisition} details data acquisition)}
    \label{fig:overall_system}
    \end{center}
    \vspace{-2em}
\end{figure}

Our proposed solution aims to support medical practitioners in their TTE workflow by introducing a framework capable of automatically classifying and grading US cardiac images. This framework aims to enhance efficiency of cardiac US scanning and reduce errors. An overview of the proposed framework is illustrated in Figure \ref{fig:overall_system}, which consists of three main components: \textbf{(1)} Data Acquisition, \textbf{(2)} AI Framework, and \textbf{(3)} TL for model adaptability. 
The image acquisition process was elaborated in detail in section \ref{subsubsec: Data Acquisition}. The AI framework is designed to train a model that combines the classification of the cardiac view and the assessment of the image quality. Here, a classification model is first trained on the CACTUS dataset, which is then extended through TL to account for image quality assessment as well. This helps reduce the memory used to load the model, as well as the time required to obtain predictions, since both tasks are managed by one DL model. Thus, our framework is able to guarantee faster classification and quality assessment results to medical practitioners. As part of our framework, new images may be submitted by medical practitioners, showcasing either new or existing cardiac views. Through this continuous flow of data, the system is continuously fine-tuned through another step of TL using the new data, ensuring its improvement, robustness, and adaptability over time. 

\subsection{Classification \& Grading Cardiac US Images}
\label{subsec: proposed DL framework}

DL models, particularly CNNs, have demonstrated remarkable performance in various computer vision tasks such as image classification or segmentation. Recently, there has been a growing interest in utilizing CNN architectures for the classification of US medical images \cite{sudharson2020ensemble,wu2018deep,zhu2021thyroid}. Different CNN architectures can be found in the literature, each varying in depth, convolutional intricacies, and optimization robustness.

In this study, we propose a novel framework tailored for classifying US images based on their cardiac views and assessing their quality based on established grade scales by medical professionals. The input of the proposed framework is the data obtained from the CAE Blue Phantom scan, detailed in section \ref{subsubsec: Data Acquisition}. Using the proposed framework, the data are used to train ResNet18 \cite{DBLP:conf/cvpr/HeZRS16}, a prominent CNN architecture in the field of computer vision, especially in the context of US images. ResNet18 comprises several convolutional layers and introduces an innovative residual connection mechanism to DL models. We opt for this architecture to initiate our experiments due to its ability to create a balance between optimization simplicity and rapid convergence.

The proposed framework comprises two primary components: a classification component and a grading component. The classification component is based on the ResNet18 architecture to categorize US images based on their cardiac views, each possessing unique characteristics and structures. Through convolutional layers, the model extracts features from the images using various filter sizes, which are then fed into fully connected layers to predict the class to which each image belongs. The CNN architecture is trained on the collected dataset with the aim of maximizing the classification accuracy of all the different classes. The grading component assigns a grade to each US image using a regression model. The grading model learns the correlation between the features of the US images and their assigned grades, which is extrapolated to predict grades for new images. 

\subsection{Transfer Learning}
\label{subsec: transfer_learning}
\begin{figure}[htp]
    \begin{center}
    \centering
\includegraphics[width=0.7\linewidth]
{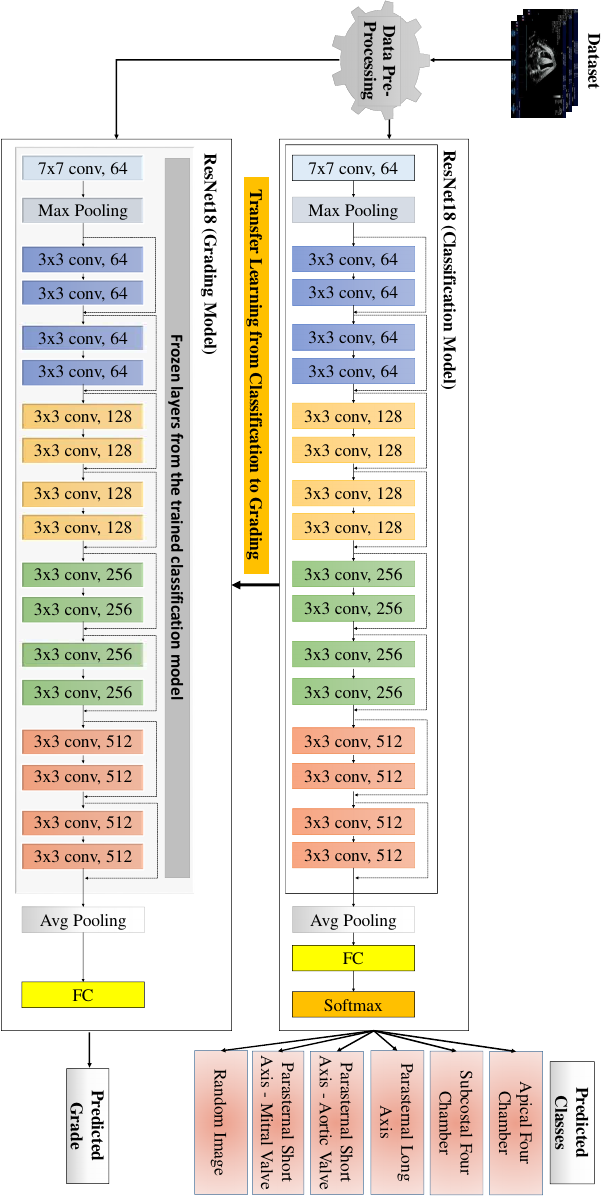}
   \caption{The training process of the classification and grading of cardiac views}
    \label{fig:trained_framework}
    \end{center}
\end{figure}

Since both the classification and grading models are expected to handle images with similar features, and to mitigate computational complexities associated with the use of multiple DL models, we propose a TL-based approach which aims to re-utilize parts of the classification model to create a grading model responsible for evaluating images. More explicitly, after training the classification model, we maintain the initial layers of the model. These layers represent the encoder (feature extractor) responsible for extracting different levels of features from the input images. On top of this feature extractor, we add a feed forward layer to predict the grades of US images and minimize grading loss. During the training of the grading model, the pre-trained encoder is frozen and its weights are not updated. This ensures the preservation of the feature extractor of the model while only training the feed forward layer responsible for predicting grades. The training process is illustrated in Figure \ref{fig:trained_framework}. 

\begin{figure}[htp]
    \begin{center}
    \centering
\includegraphics[width=0.92\linewidth]{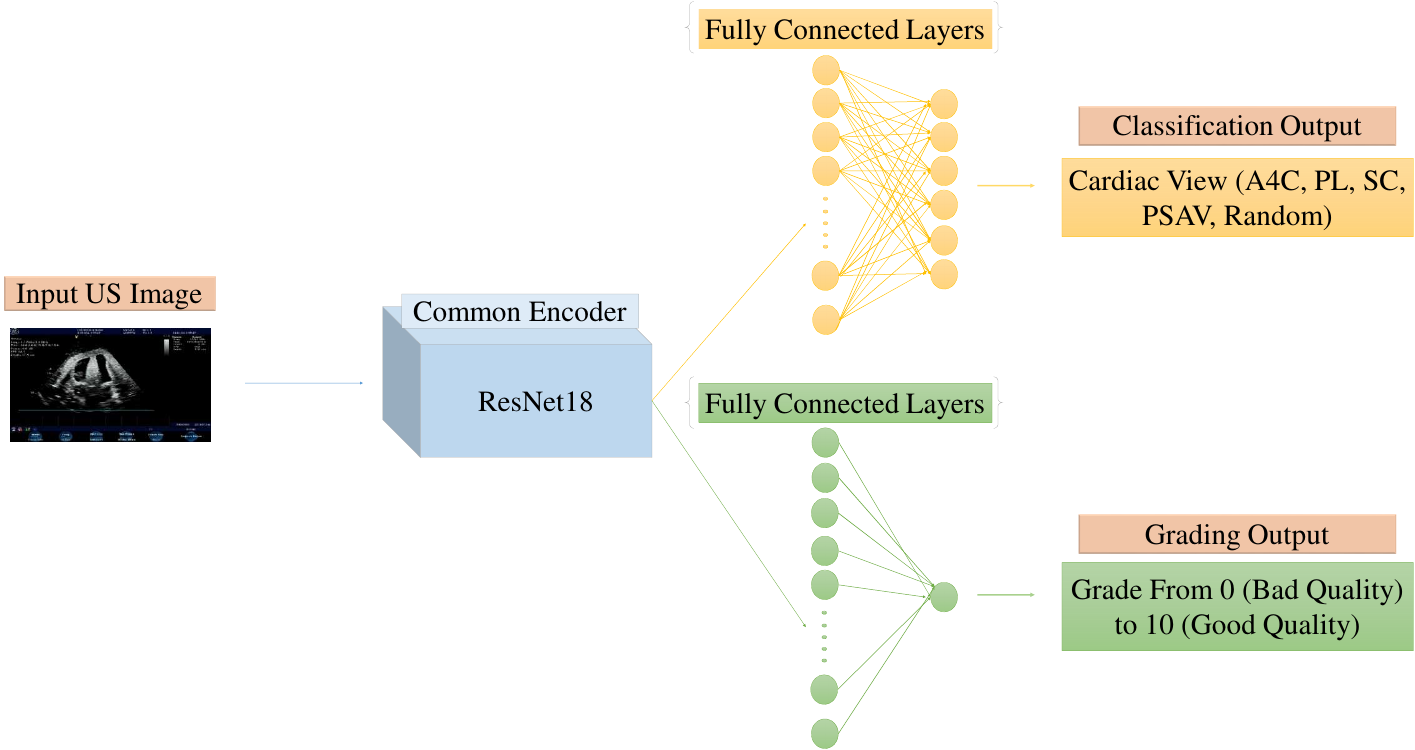}
    \caption{The architecture of the final proposed AI framework built upon ResNet18}
    \label{fig:common_encod}
    \end{center}
    \vspace{-1.5em}
\end{figure}

By adopting this strategy, our aim is to decrease computation time and training parameters, as well as Floating Point Operations (FLOPs) through utilizing a common encoder and different heads, as opposed to employing two separate ResNet18 models. Following the training process, the final outcome is one model with a common encoder based on ResNet18 and two different heads, one for classification and one for grading, which can produce outputs in parallel, as depicted in Figure \ref{fig:common_encod}. To evaluate the effectiveness of TL, we compared its performance with the multi-task learning (MTL) approach \cite{zhang2021survey}, which involves creating a single encoder model with two heads: one for classification and one for grading. This design allows for simultaneous training of both tasks using combined loss function, whereas TL typically involves sequential fine-tuning of the model on the specific task. Their performance will be analyzed and discussed in section \ref{subsec: Results and Discussion}.

\begin{figure}[H]
    \begin{center}
\includegraphics[width=0.8\linewidth]
{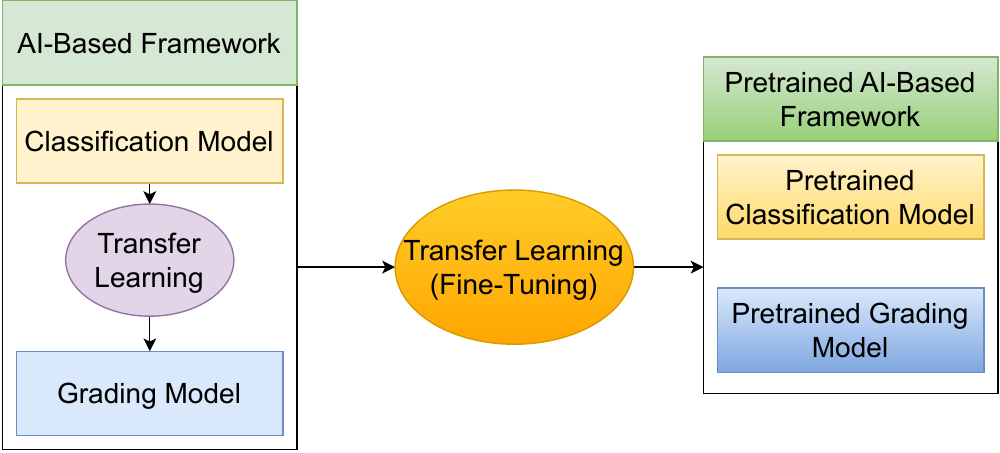}
    \caption{The process of application of transfer learning for the proposed framework}
    \label{fig:process_tl}
    \end{center}
    \vspace{-2em}
\end{figure}

As part of the presented framework, we also utilize TL again to improve the model when possible. This is achieved by fine-tuning the model with new US images, if available. Initially, the framework is trained and validated on the initial dataset, which includes four cardiac views (A4C, PL, PSAV, SC), along with random images. Subsequently, once the model is trained, the new images representing PSMV are incorporated, and TL is used to fine-tune the pretrained model and extend it to include the classification and grading of the new view, along with the previous ones. This step is essential because the model must be adaptable to accept new US cardiac images, as US machines and scanning angles may vary. Thus, the model should always be updatable, and TL facilitates achieving this goal. Figure \ref{fig:process_tl} describes the process of application of TL for fine-tuning the model in the proposed framework.

\section{Experiments}
\label{sec: Experiments}
This section presents the evaluation of the proposed AI framework, which is primarily designed to assist medical professionals in the classification and grading of cardiac US images. The framework's performance on the CACTUS dataset was assessed using multiple criteria. The classification model is evaluated using metrics such as accuracy, F1 score, precision, and confusion matrix. The grading model is additionally evaluated using the loss value between the actual grades and the predicted values. We also evaluate the performance of TL in terms of the model's adaptability to new data. Additionally, the proposed ResNet18 architecture is compared with other CNN architectures, including ResNet50 \cite{DBLP:conf/cvpr/HeZRS16}, VGGNet19 \cite{simonyan2014very}, InceptionNet v3 \cite{szegedy2016rethinking}, and AlexNet \cite{krizhevsky2012imagenet}. Finally, the effectiveness of the AI framework in handling real-time scans is assessed by presenting statistical results from the scans and using a questionnaire to gather feedback, with a particular focus on responsiveness, usefulness, and intuitiveness.

\subsection{Experiments Setup}
\label{subsec: Experiments Setup}


Before the training process, the collected data undergoes common image preprocessing techniques to ensure uniformity and stability during learning. The images are first cropped to remove irrelevant portions, such as machine parameters and settings, focusing only on the cardiac view. A resizing step is then applied to ensure that all images are of a consistent size of 224x224 pixels. Following this, the images are normalized to standardize pixel values, improving the model’s convergence during training. Finally, the dataset is divided into three subsets: 70\% for training, 10\% for validation, and 20\% for testing. Regarding the experiment parameters and setup, we utilize hyperparameter tuning to optimize the performance of our framework. This involves adjusting parameters such as the learning rate, epoch, and batch size until getting satisfactory training results. After experimentation, the best performance was achieved with 30 epochs, a batch size of 128, and a learning rate of 0.001. In this study, stochastic gradient descent serves as the optimizer for both classification and grading scenarios. The categorical cross-entropy (CCE) is employed for training the classification model, while mean squared error (MSE) is utilized as the objective function for training the grading model. All experiments are conducted using PyTorch libraries and performed within the Compute Canada platform, using the CEDAR remote machine, which offers 100,400 CPU cores and 1,352 GPU devices for computation. Specifically, a configuration was used with one GPU, 4GB of memory per CPU, utilizing a single node and running eight tasks concurrently.

\subsection{Results and Discussion}
\label{subsec: Results and Discussion}
In the upcoming sections, we explore various approaches to assess the proposed framework. Initially, we analyze the performance of the classification model through several metrics such as accuracy, F1 score, precision, recall and confusion matrix, as well as the grading model using its loss plot. In this work, the terms `error' and `loss' are used interchangeably to refer to the result of the objective function MSE used during the training of the grading model. The framework is additionally compared with other state-of-the-art CNN architectures. We also assess TL on new data to evaluate the adaptability of the framework. Subsequently, we analyze the impact of TL from the classification model to the grading model, illustrating improvements in optimization of prediction time, training parameters, and FLOPs.  Lastly, we describe the real-time evaluation of the proposed framework which was assessed by cardiac imaging experts.

\subsubsection{Performance of the Classification and Grading Models}
\label{subsubsec: Performance of classification and grading}
\begin{figure}[h]
    \begin{center}
    \centering
\includegraphics[width=0.9\linewidth]{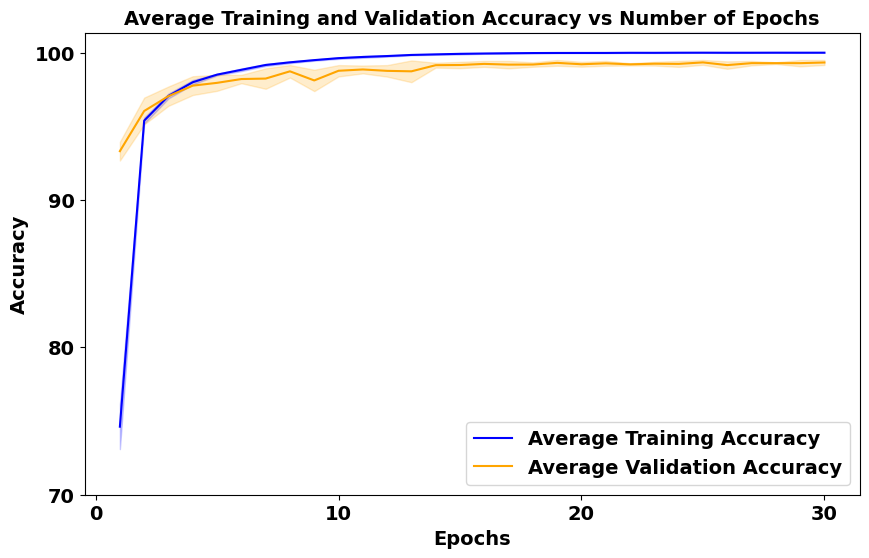}
    \caption{Classification accuracy plot of the proposed framework}
\label{fig:classification_results_accuracy_loss}
    \end{center}
    \vspace{-1em}
\end{figure}

\begin{figure}[h]
    \begin{center}
    \centering
\includegraphics[width=0.9\linewidth]{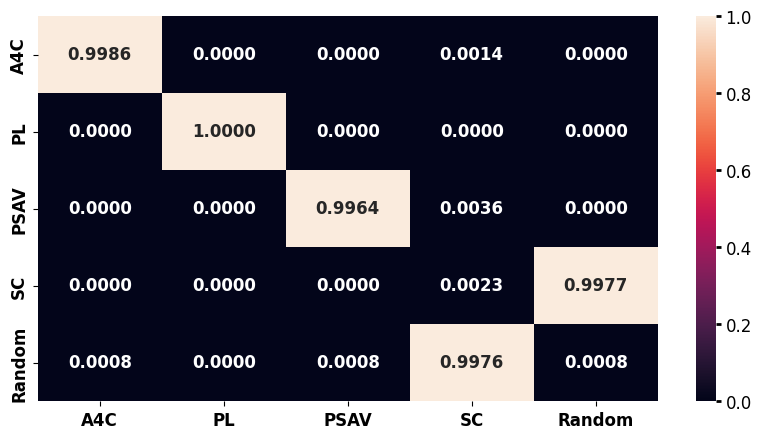}
    \caption{Confusion matrix of the proposed framework}
    \label{fig:confusion}
    \end{center}
    \vspace{-2em}
\end{figure}

Training a ResNet18 model for cardiac view classification produces significant results. Figure \ref{fig:classification_results_accuracy_loss} shows the training and validation accuracies of the classification model, resulting from averaging five times different training runs. From this plot, it is evident that the model converges around the seventh epoch, achieving an accuracy of 99.97\%. 
Similarly, the validation curve converges to similar values, with an accuracy of 99.43\%, indicating the model's ability to generalize to unseen data. Additionally, both Table \ref{tab:classification} and the corresponding confusion matrix in Figure \ref{fig:confusion} highlight the remarkable performance of ResNet18 on the validation set, accurately classifying all classes.

\begin{table}[h]
  \centering
  \begin{tabular}{|c |c|c|c|}
\hline
    Class & Precision  & Recall & F1 Score \\ \hline
A4C & 1.0 & 1.0 & 1.0 \\
\hline

PL & 1.0  & 1.0  & 1.0 \\
\hline

PSAV & 1.0  & 1.0  & 1.0 \\
\hline

SC & 1.0 & 1.0 & 1.0 \\
\hline

Random Images & 0.99 & 1.0 & 1.0 \\
\hline

\end{tabular}
  \caption{Classification report of the proposed framework}
\label{tab:classification}
\vspace{-1em}
\end{table}

\begin{figure}[h]
    \begin{center}
    \centering
\includegraphics[width=0.9\linewidth]{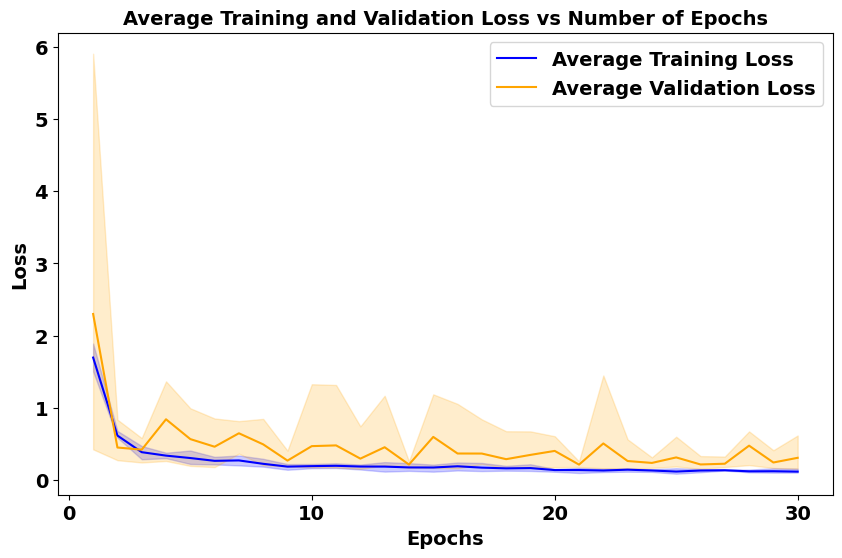}
    \caption{Grading results of the proposed framework}
    \label{fig:regression_results}
    \end{center}
    \vspace{-2em}
\end{figure}

Similarly, the grading model, derived from TL using the classification-based ResNet18 model, demonstrates satisfactory performance. This is illustrated in Figure \ref{fig:regression_results}, which showcases the training and validation performance of the model, resulting from averaging five training runs. The grading model begins to converge around epoch 10, reaching in average a training loss of 0.1154. Although the validation loss exhibits slightly higher values, converging in average to 0.3067 after epoch 30, this is considered normal given the complexity of the data. Nonetheless, these results remain helpful and informative for medical practitioners, as the error of the model is nearly 0.4, which means that, on average, the model predicts grades within a range of ±0.4 which still provides a good indication of the quality of the image.

\begin{figure}[h]
  \begin{subfigure}{.49\textwidth}
  \centering
    \includegraphics[width=1\linewidth]{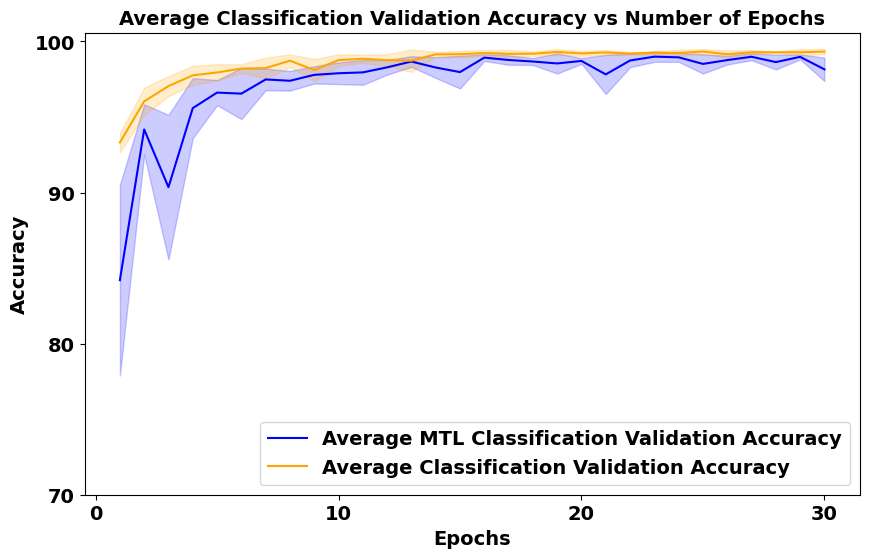}
  \end{subfigure}%
  \begin{subfigure}{.49\textwidth}
  \centering
    \includegraphics[width=1\linewidth]{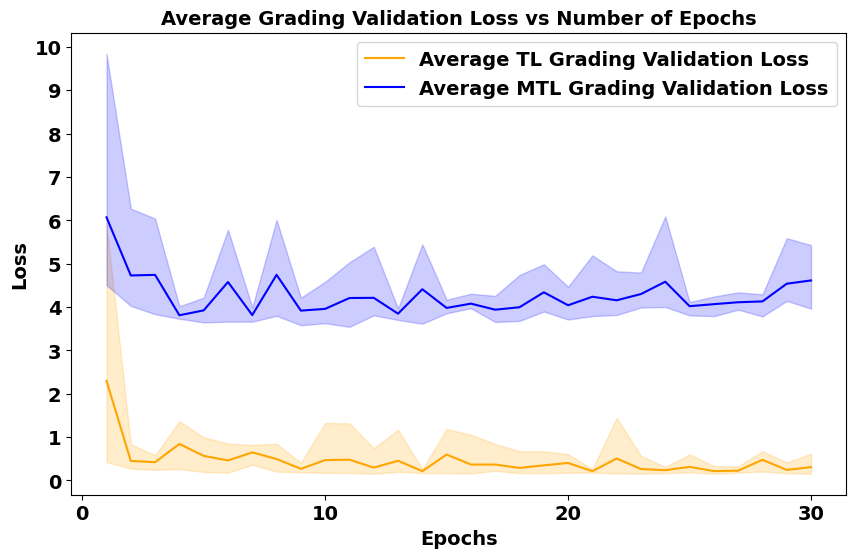}
  \end{subfigure}%
  \caption{Performance comparison between transfer learning and multi-task learning}
  \label{fig:MTL_vs_TL}
  \vspace{-1em}
\end{figure}

The performance of TL is also compared to MTL to determine which approach is better suited for this scenario. In MTL, the classification and grading tasks are trained simultaneously using a combined loss function.
The two graphs in Figure \ref{fig:MTL_vs_TL} display the performance of TL versus MTL for both classification and grading tasks, averaged over five training runs. The plots show that classification performance remains comparable between the two approaches. However, for the grading task, TL significantly outperforms MTL. This superior performance of TL can be attributed to the fact that during TL, the grading model benefits from the feature representations learned during the classification pre-training phase. In contrast, while MTL allows for simultaneous training of both tasks, it may not be able to fully optimize the loss function of both tasks at the same time due to the use of a combined loss function. The joint learning of two tasks in MTL can lead to competing gradients, making it more difficult for the model to specialize and optimize for each task individually, especially for more complex tasks like grading.

There are several CNN architectures that differ in terms of depth, convolutional sizes, and computational capacity. In the scope of this study, we initially evaluate the performance of ResNet18 and designate it as the foundational element of our framework. However, recognizing the need for a more comprehensive assessment, we conduct a comparative analysis to evaluate other state-of-the-art models prevalent in literature, including VGGNet19 \cite{simonyan2014very}, InceptionNet v3 \cite{szegedy2016rethinking} and AlexNet \cite{krizhevsky2012imagenet}. AlexNet is chosen for its pioneering contribution to the development of CNNs, while VGGNet19 is selected to assess the impact of its deep architecture and small convolutional filters on the classification and grading of cardiac US images. InceptionNet v3 is included for its use of multi-scale convolution filters, which allow the model to capture complex features at various scales by widening the network rather than increasing its depth. Additionally, we examine ResNet50 \cite{DBLP:conf/cvpr/HeZRS16} to evaluate the impact of deeper models on this dataset. This comparison aims to determine which architecture achieves optimal performance in terms of both image classification and grading.

To compare the CNN architectures, we adhere to the identical training setup. Table \ref{tab:classification_ablation} displays the classification performance of each model in terms of accuracy, precision, and F1 score metrics.  
According to the results presented in Table \ref{tab:classification_ablation}, ResNet18 alongside ResNet50 and InceptionNet v3, emerged as the top-performing models, showing an accuracy of 99.0\%. VGGNet19 also yielded exceptionally high accuracies. These findings were further supported by metrics such as precision and F1 score, which reinforced the superior performance of these models. In contrast, ResNet18 (when applied in the context of MTL), along with AlexNet, exhibited the lowest testing accuracy at 98\%.

\begin{table}[ht!]
  \centering
  \begin{tabular}{|c |c|c|c|}
\hline
    Model & Accuracy (\%)  & Precision & F1 Score\\ \hline
VGGNet19 & 99.0 & 0.99 &  0.99\\
\hline

AlexNet & 98.0  & 0.98 & 0.98\\
\hline

InceptionNet v3 & 99.0 & 0.99 &  0.99\\
\hline

ResNet50 & 99.0 & 0.99 &  0.99\\
\hline

ResNet18 &  100 & 1.00 &  1.00 \\
\hline

ResNet18 (MTL) &  98.81 & 0.98 & 0.98 \\

\hline
\end{tabular}
  \caption{Classification results for different models}
\label{tab:classification_ablation}
\vspace{-1em}
\end{table}

      



The grading model, obtained through TL from the classification-based ResNet18 model, shows the best test performance in the grading task, outperforming ResNet18 without TL, as shown in Table \ref{tab:regression}, along with models like AlexNet, InceptionNet v3, and ResNet50. Conversely, when using the MTL approach, ResNet18 exhibits the lowest performance, with a significantly higher test loss compared to other models. This highlights the fact that TL is a better choice for grading tasks than MTL. Consequently, while utilizing TL from classification leads to improvements in computation and memory efficiency, it comes at the expense of validation and test performance. To address these limitations, we plan to enrich the CACTUS dataset with a wider range of image grade variations to enhance the robustness of the proposed framework.

\begin{table}[h]
  \centering
  \begin{tabular}{|c |c|}
\hline
    Model & Test Loss \\ \hline
VGGNet19 &  0.2185\\
\hline

AlexNet & 0.1492  \\
\hline

InceptionNet v3 & 0.1803 \\
\hline

ResNet18 &  0.2485\\
\hline

ResNet50 &  0.1670\\
\hline

ResNet18 (TL from classification) &  0.1077\\
\hline
ResNet18 (MTL) &  9.916\\
\hline

\end{tabular}
  \caption{Grading results for different models}
\label{tab:regression}
\vspace{-1em}
\end{table}

\begin{figure}[h]
    \begin{center}
    \centering
\includegraphics[width=0.9\linewidth]{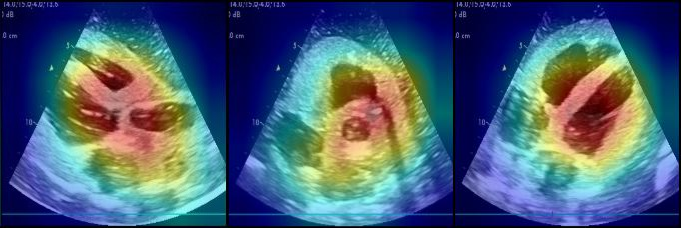}
    \caption{Grad-CAM++ visualizations for the classification model: PL (left), PSAV (middle), and SC (right)}
    \label{fig:gc_classification}
    \end{center}
    \vspace{-1em}
\end{figure}

\begin{figure}[h]
    \begin{center}
    \centering
\includegraphics[width=0.9\linewidth]{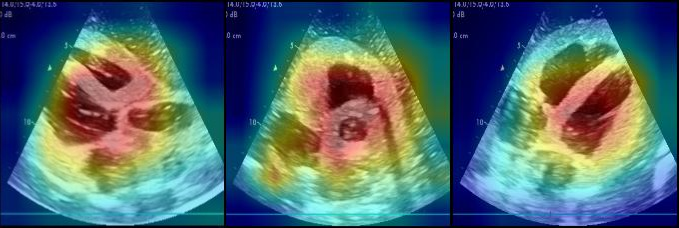}
    \caption{Grad-CAM++ visualizations for the grading model: PL (left), PSAV (middle), and SC (right)}
    \label{fig:gc-grading}
    \end{center}
    \vspace{-2em}
\end{figure}

An additional key aspect to consider is the interpretability of the proposed AI framework. Explainable artificial intelligence (XAI) is an emerging field focused on making DL models interpretable, enabling users to understand the reasons behind specific predictions by highlighting the relevant features in the input image that contributed to the model's output \cite{suara2023grad}. In the medical field, comprehending the behavior of DL models is essential, as it directly influences clinical decisions and outcomes. In this context, our work focuses on gaining insights into how the proposed AI framework performs both classification and grading tasks, offering a specific case for understanding model behavior in medical applications. To achieve this, we have selected Gradient-weighted Class Activation Mapping (Grad-CAM), one of the most widely used methods for explainable DL, with its variant Grad-CAM++ \cite{selvaraju2017grad,chattopadhay2018grad}. The results are presented in Figures \ref{fig:gc_classification} and \ref{fig:gc-grading}. In the Grad-CAM++ heatmap, different colors represent the intensity of focus by the model. Red indicates the areas the model considers most important, while yellow and blue represent decreasing levels of attention. From these visualizations, we observe that the AI framework predominantly focuses on the central region of the image (highlighted in red), which corresponds to the identification of the four chambers in the PL US cardiac image. Similarly, in the PSAV and SC cardiac views, the model focuses on the areas where the structure of the view is clearly delineated. This suggests that the model prioritizes this cardiac structural features for both classification and grading tasks.

\subsubsection{Adaptability of the Proposed Framework to New Data}
\label{subsubsec: adaptability analysis}
\begin{figure}[H]
    \begin{center}
    \centering
\includegraphics[width=0.9\linewidth]{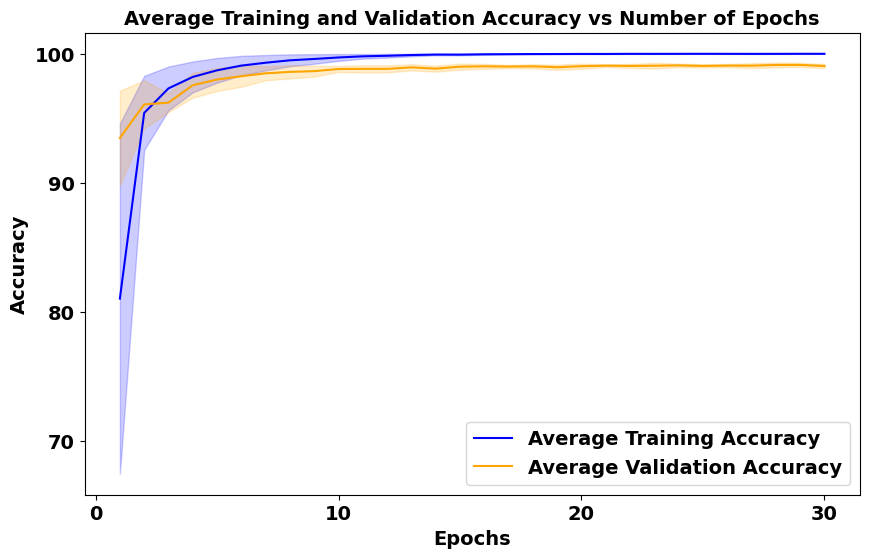}
    \caption{Fine-tuning classification accuracy plots}
\label{fig:classification_tl_results_accuracy_loss}
    \end{center}
    \vspace{-2em}
\end{figure}

In this section, we examine the adaptability of our framework when introducing a new view, i.e. ``PSMV", to the initial dataset, thereby expanding the dataset to include six classes. We then fine-tune the pre-trained classification and grading models on this new dataset. Regarding the performance of the classification model trained on the six classes, the analysis of Figure \ref{fig:classification_tl_results_accuracy_loss} reveals, based on the average of five runs, that the model converges rapidly for both training and validation scenarios, reaching an accuracy of 99.99\%.




\begin{figure}
    \begin{center}
    \centering
\includegraphics[width=0.9\linewidth]{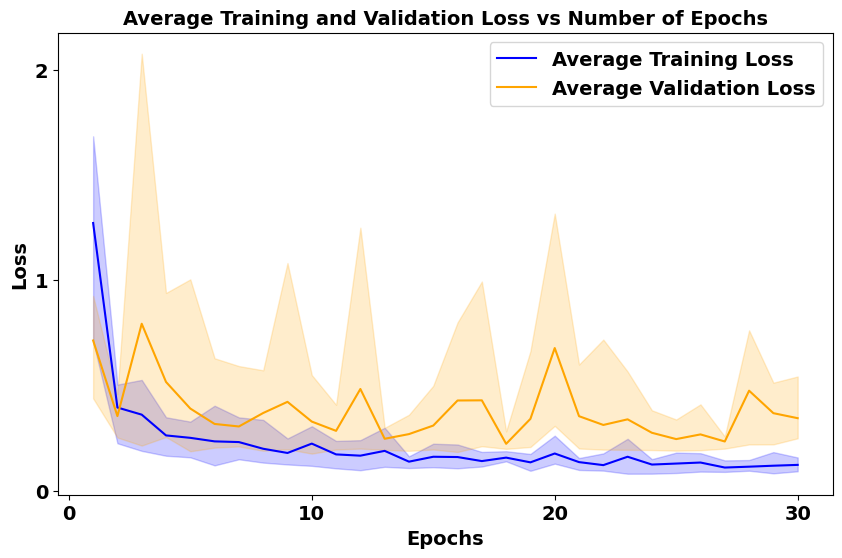}
    \caption{Fine-tuning grading loss plot}
    \label{fig:loss_ps3}
    \end{center}
    \vspace{-2em}
\end{figure}

Similarly, the grading model produces promising results. This is evident from Figure \ref{fig:loss_ps3}, which illustrates the training and validation curves. The model begins to converge around epoch 21, reaching a training loss of 0.091 and validation loss of 0.542.

\subsubsection{Computation Analysis}
\label{subsubsec: Computation analysis}
In this section, we explore the impact of employing TL to obtain the grading model within our framework. According to the data presented in Table \ref{tab:computation_analysis}, the grading model derived through TL from the classification model demonstrates a notably smaller number of trainable parameters compared to a model trained entirely from scratch. This indicates that employing the classification and grading models with a shared encoder (through TL) consumes less computational memory and requires less prediction time than using two separate models for classification and grading. Furthermore, the performance of TL is also compared with MTL. According to the results in the table, it is evident that TL and MTL have similar prediction time, which is attributed to the fact that they have similar trainable parameters and similar FLOPs. This is due to the fact that the final model for both methods, TL and MTL, is nearly the same with one encoder backbone and two output heads. Both methods result in a final model with 1.18 GFLOPs. This outperforms the traditional method of using two completely separate models for classification and grading, which would result in double the trainable parameters and FLOPs count, requiring nearly 3.6 GFLOPs. While both TL and MTL give similar results with regards to computational complexity, it was previously shown in Section \ref{subsubsec: Performance of classification and grading} that TL outperforms MTL in its learning capabilities, making it the favored approach.

\begin{table}[H]
  \centering
  \begin{tabular}{|c|p{3 cm}|c|}
\hline
    Model & Prediction Time  & Trainable Parameters \\ \hline
Grading Model (without TL) & \centering 3.00 ms  & 11177025\\
\hline

Grading Model (with TL) & \centering 2.90 ms & 513 \\
\hline

Classification Model& \centering  2.99 ms & 11179590\\
\hline

MTL Model& \centering  3.27 ms & 11180103 \\
\hline

\end{tabular}
  \caption{Computation analysis}
\label{tab:computation_analysis}
\vspace{-1em}
\end{table}

\subsubsection{Real-time Evaluation of the Proposed Framework}
\label{subsubsec: Questionnaire}

\begin{figure}[H]
    \begin{center}
    \centering
\includegraphics[width=1.0\linewidth]{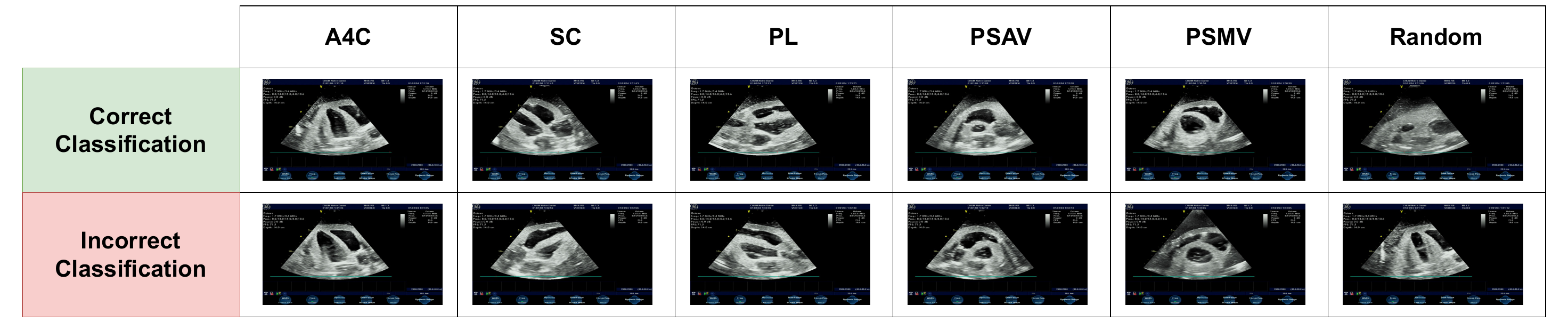}
    \caption{Real time scans: examples of classification results}
    \label{real_time_class}
    \end{center}
    \vspace{-1em}
\end{figure}

To evaluate the proposed AI framework, we also conducted real-time scans, providing the framework with real-time cardiac US images. For these real-time experiments, we kept the same US parameters for all views. We obtained the results from the AI model and compared them with expert feedback in terms of classification accuracy and grading loss. Based on the results, we observed that all cardiac views were accurately classified, with classification accuracy reaching up to 96\%, and with rare challenges faced for the Random and PSAV classes. For the Random class, there were few scenarios where the model misclassified a view as Random due to its low quality. While an expert can identify the cardiac view (even if it is incomplete), the model classifies it as random due to the low quality. As for the PSAV class, it was challenging for the model to detect it in certain scenarios due to the absence of dynamic valves and the static nature of the cardiac phantom, making it difficult sometimes to be differentiated from random views. Therefore, in the future, acquiring a phantom with dynamic valves would be a valuable approach, enabling more precise differentiation between the PSAV view and other views. Figure \ref{real_time_class} shows the classification results during real-time scans, including both correct classification results and cases of misclassification. As generally shown in the figure, misclassifications are typically incorrectly classified as Random. An image that should be classified as Random is sometimes classified as A4C, which is due to the nature of the phantom that does not allow for precise differentiation between the A4C class and the Random class. The same issue applies to the PSAV class.

\begin{figure}[H]
    \begin{center}
    \centering
\includegraphics[width=0.8\linewidth]{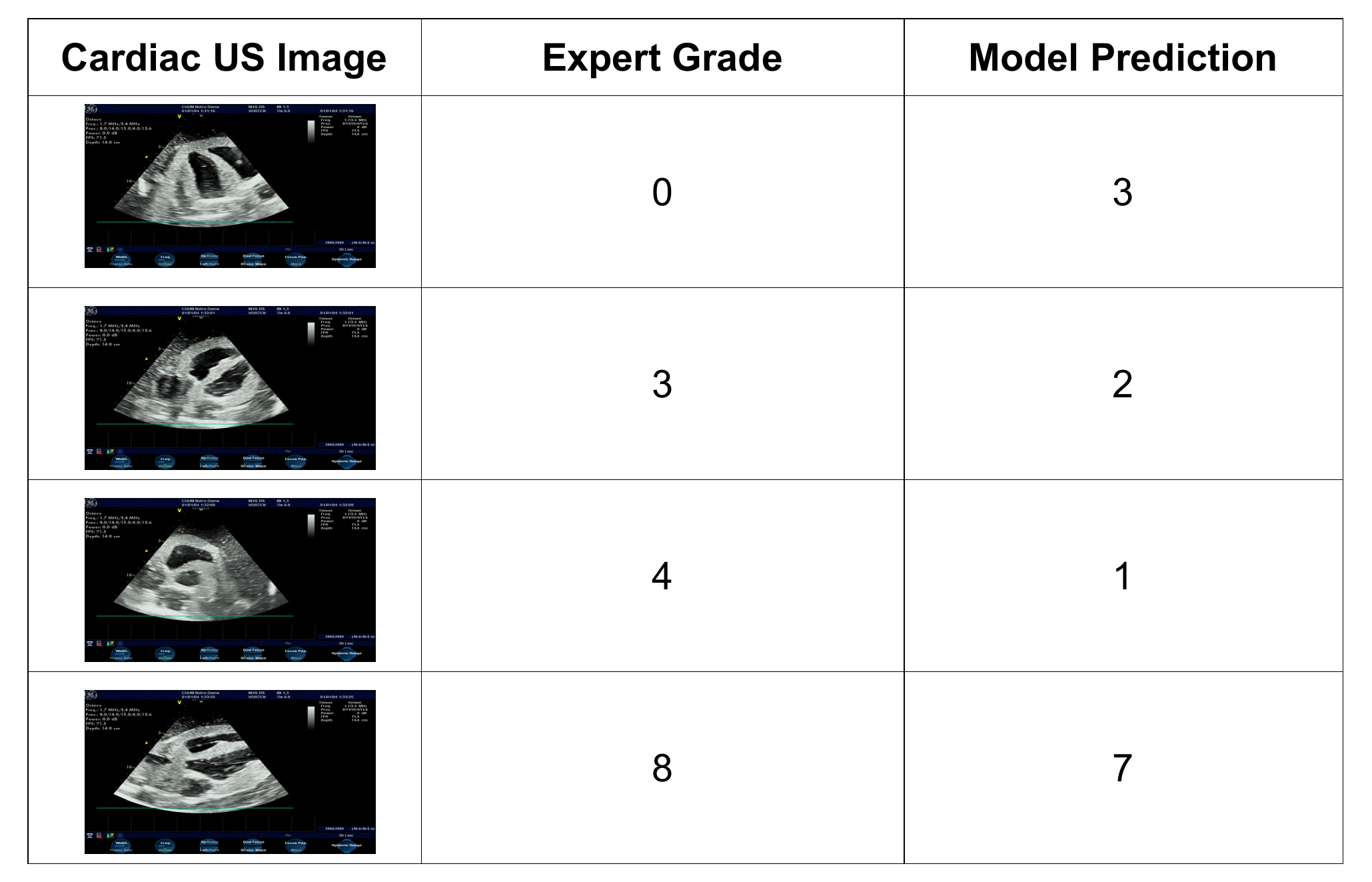}
    \caption{Real time scans: examples of misgraded results}
    \label{real_time_grading}
    \end{center}
    \vspace{-1em}
\end{figure}

On the other hand, we also obtained grading results from the AI model and compared them with expert assessments. Figure \ref{real_time_grading} illustrates the grading outcomes during real-time scans, including the results of incorrect grading when compared to expert assessments. The results were also promising, with the cardiac views SC, PL, and Random showing better performance, while A4C and PSAV showed less satisfactory results. The factors influencing misgrading include:
\begin{itemize}
\item  The static nature of the CAE Blue Phantom heart and the absence of structures like valves, which makes it difficult to distinguish views like PSAV and A4C.
\item  Multiple iterations are necessary to grade the ground truth images in order to thoroughly review the entire dataset and establish grading based on the overall quality of the dataset.
\item The differences between successive grades are very subtle.
\end{itemize}

At the end of the experiments, we also provided the two cardiac experts with the questionnaire to assess the framework's performance and verify if it can assist medical practitioners during US cardiac scanning. The questionnaire aims to assess the framework's outcomes and its performance in handling real-time scans in terms of responsiveness, usefulness, and intuitiveness of results. It includes nine questions that are described in the Table \ref{tab:questionnaire} below. The column ``Answer" describes the type of answers accepted for each question.  

\begin{table}[H]
  \centering
  \begin{tabular}{|p{10cm}|c|}
\hline
   \centering Question & Answer \\ \hline
On a scale of 0 to 10, How accurately does the framework categorize the scanned ultrasound cardiac images? & Scale (0 - 10)  \\
\hline

Could you please explain what aspects of the results are missing if the classification is not accurate? & Descriptive Text\\
\hline

On a scale of 0 to 10, how confident is the model in predicting the grading outcome? &  Scale (0 - 10)\\
\hline

If the outcomes are incorrect, according to you, what makes the model's predictions low? & Descriptive Text\\
\hline

On a scale of 0 to 10, how would you rate the responsiveness of the models' real-time simulation? &  Scale (0 - 10)\\
\hline

On a scale of 0 to 10, how would you rate the usefulness of the results of the model? &  Scale (0 - 10)\\
\hline

On a scale of 0 to 10, how would you rate the intuitiveness of the results? & Scale (0 - 10)\\
\hline

On a scale of 0 to 10, how would you rate the overall experience you have had with framework? & Scale (0 - 10)\\
\hline

What suggestions do you have for additional elements to incorporate into the framework to assist medical practitioners during ultrasound cardiac scanning examinations? & Descriptive Text\\
\hline

\end{tabular}
  \caption{Questionnaire for performance analysis by medical experts}
\label{tab:questionnaire}
\vspace{-1em}
\end{table}

\begin{figure}
    \begin{center}
    \centering
\includegraphics[width=1.0\linewidth]{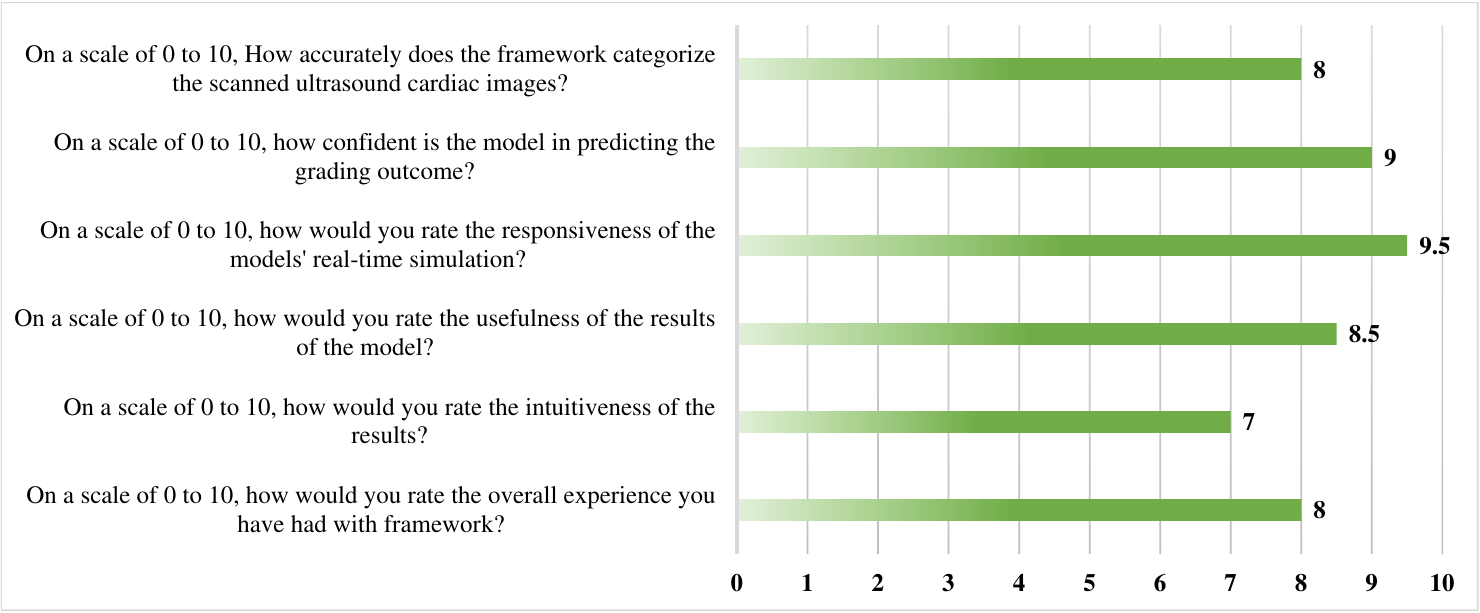}
    \caption{Average scores for the questionnaire}
    \label{fig:questionnaire_results}
    \end{center}
    \vspace{-1em}
\end{figure}

Based on the answers of the cardiac experts, we received the following feedback, as depicted in Figure \ref{fig:questionnaire_results}, which shows the average ratings on a scale of 0 to 10.
From the graph depicted in Figure \ref{fig:questionnaire_results}, we understand that the model achieves satisfactory results in terms of classification accuracy and also in terms of confidence in predicting grades of US images. The framework also shows good performance in terms of responsiveness and relevance of results, even though it requires more effort to make the results more intuitive. Overall, the cardiac experts rated their experience with our solution well, achieving an average score of 8 out of 10.

As part of their answers to the questionnaire, the cardiac practitioners also proposed new features that could be included in the model, such as optimizing US parameters to better visualize cardiac windows, addressing the amount of random views classification, and choosing a better acoustic window to reduce the scanning time.

\section{Conclusion}
\label{sec: concolusions}
With the aim of aiding medical professionals during TTE examinations, this paper presents an AI-driven framework to assist in classifying and grading cardiac US images. Given the scarcity of literature on this topic due to limited public datasets, our paper presents the first openly graded dataset named CACTUS, incorporating various categories of cardiac views, exceeding the conventional cardiac views typically found in literature. These images originate from scanning the CAE Blue Phantom and are graded on a scale from 0 (poor quality) to 10 (highest quality). Moreover, the paper introduces a framework utilizing the ResNet18 architecture and it comprises two primary blocks: one dedicated to classifying views and the other formed through TL to assess the quality of cardiac images by providing grades. The AI framework shows good results with a classification validation accuracy of 99.43\% and a grade prediction loss converging to 0.3067. To showcase its robustness, we further fine-tune the framework using new images representing additional cardiac views, yielding satisfactory outcomes with a classification validation accuracy of 99.99\% and minimal grading validation loss of 0.542.

For future work, we plan to expand the dataset to encompass a broader range of cardiac anatomies from diverse phantoms. Additionally, we aim to extend our approach to real-time analysis of dynamic human heart scans, moving beyond static cardiac phantom imaging. This transition will facilitate new research directions, including numerical quantification of cardiac function and real-time assessment of heart dynamics. While TL has demonstrated strong potential in our study, certain limitations remain to be addressed in future work. TL often relies on human expertise for guiding the transfer process, necessitating the development of more efficient integration methods to reduce computational costs. Additionally, the risk of negative transfer, where knowledge from a source domain hinders rather than enhances performance in the target domain, remains a crucial issue, particularly when dealing with domain shifts. To mitigate domain shifts, we plan to employ domain adaptation techniques, such as adversarial training and self-supervised learning, to improve the model's ability to generalize across different datasets and imaging conditions.

\bibliographystyle{model1-num-names}
\bibliography{references.bib}

\begin{thebibliography}{47}
\expandafter\ifx\csname natexlab\endcsname\relax\def\natexlab#1{#1}\fi
\providecommand{\bibinfo}[2]{#2}
\ifx\xfnm\relax \def\xfnm[#1]{\unskip,\space#1}\fi
\bibitem[{Javaid et~al.(2022)Javaid, Haleem, Singh, Suman, and Rab}]{javaid2022significance}
\bibinfo{author}{M.~Javaid}, \bibinfo{author}{A.~Haleem}, \bibinfo{author}{R.~P. Singh}, \bibinfo{author}{R.~Suman}, \bibinfo{author}{S.~Rab},
\newblock \bibinfo{title}{Significance of machine learning in healthcare: Features, pillars and applications},
\newblock \bibinfo{journal}{International Journal of Intelligent Networks} \bibinfo{volume}{3} (\bibinfo{year}{2022}) \bibinfo{pages}{58--73}.
\bibitem[{May(2021)}]{may2021eight}
\bibinfo{author}{M.~May},
\newblock \bibinfo{title}{Eight ways machine learning is assisting medicine},
\newblock \bibinfo{journal}{Nat Med} \bibinfo{volume}{27} (\bibinfo{year}{2021}) \bibinfo{pages}{2--3}.
\bibitem[{Kumari and Singh(2023)}]{kumari2023deep}
\bibinfo{author}{S.~Kumari}, \bibinfo{author}{P.~Singh},
\newblock \bibinfo{title}{Deep learning for unsupervised domain adaptation in medical imaging: Recent advancements and future perspectives},
\newblock \bibinfo{journal}{Computers in Biology and Medicine}  (\bibinfo{year}{2023}) \bibinfo{pages}{107912}.
\bibitem[{Averbuch et~al.(2022)Averbuch, Sullivan, Sauer, Mamas, Voors, Gale, Metra, Ravindra, and Van~Spall}]{averbuch2022applications}
\bibinfo{author}{T.~Averbuch}, \bibinfo{author}{K.~Sullivan}, \bibinfo{author}{A.~Sauer}, \bibinfo{author}{M.~A. Mamas}, \bibinfo{author}{A.~A. Voors}, \bibinfo{author}{C.~P. Gale}, \bibinfo{author}{M.~Metra}, \bibinfo{author}{N.~Ravindra}, \bibinfo{author}{H.~G. Van~Spall},
\newblock \bibinfo{title}{Applications of artificial intelligence and machine learning in heart failure},
\newblock \bibinfo{journal}{European Heart Journal-Digital Health} \bibinfo{volume}{3} (\bibinfo{year}{2022}) \bibinfo{pages}{311--322}.
\bibitem[{Bhushan et~al.(2023)Bhushan, Pandit, and Garg}]{bhushan2023machine}
\bibinfo{author}{M.~Bhushan}, \bibinfo{author}{A.~Pandit}, \bibinfo{author}{A.~Garg},
\newblock \bibinfo{title}{Machine learning and deep learning techniques for the analysis of heart disease: a systematic literature review, open challenges and future directions},
\newblock \bibinfo{journal}{Artificial Intelligence Review} \bibinfo{volume}{56} (\bibinfo{year}{2023}) \bibinfo{pages}{14035--14086}.
\bibitem[{Petmezas et~al.(2024)Petmezas, Papageorgiou, Vassilikos, Pagourelias, Tsaklidis, Katsaggelos, and Maglaveras}]{petmezas2024recent}
\bibinfo{author}{G.~Petmezas}, \bibinfo{author}{V.~E. Papageorgiou}, \bibinfo{author}{V.~Vassilikos}, \bibinfo{author}{E.~Pagourelias}, \bibinfo{author}{G.~Tsaklidis}, \bibinfo{author}{A.~K. Katsaggelos}, \bibinfo{author}{N.~Maglaveras},
\newblock \bibinfo{title}{Recent advancements and applications of deep learning in heart failure: A systematic review},
\newblock \bibinfo{journal}{Computers in Biology and Medicine}  (\bibinfo{year}{2024}) \bibinfo{pages}{108557}.
\bibitem[{Sadeghpour and Alizadehasl(2018)}]{Sadeghpour2018}
\bibinfo{author}{A.~Sadeghpour}, \bibinfo{author}{A.~Alizadehasl},
\newblock \bibinfo{title}{Echocardiography},
\newblock \bibinfo{journal}{Practical Cardiology}  (\bibinfo{year}{2018}) \bibinfo{pages}{67--111}.
\bibitem[{Ciampi and Villari(2007)}]{ciampi2007role}
\bibinfo{author}{Q.~Ciampi}, \bibinfo{author}{B.~Villari},
\newblock \bibinfo{title}{Role of echocardiography in diagnosis and risk stratification in heart failure with left ventricular systolic dysfunction},
\newblock \bibinfo{journal}{Cardiovascular ultrasound} \bibinfo{volume}{5} (\bibinfo{year}{2007}) \bibinfo{pages}{1--12}.
\bibitem[{Capotosto et~al.(2018)Capotosto, Massoni, De~Sio, Ricci, Vitarelli et~al.}]{capotosto2018early}
\bibinfo{author}{L.~Capotosto}, \bibinfo{author}{F.~Massoni}, \bibinfo{author}{S.~De~Sio}, \bibinfo{author}{S.~Ricci}, \bibinfo{author}{A.~Vitarelli}, et~al.,
\newblock \bibinfo{title}{Early diagnosis of cardiovascular diseases in workers: role of standard and advanced echocardiography},
\newblock \bibinfo{journal}{BioMed Research International} \bibinfo{volume}{2018} (\bibinfo{year}{2018}).
\bibitem[{Balaji et~al.(2015)Balaji, Subashini, and Chidambaram}]{balaji2015automatic}
\bibinfo{author}{G.~Balaji}, \bibinfo{author}{T.~Subashini}, \bibinfo{author}{N.~Chidambaram},
\newblock \bibinfo{title}{Automatic classification of cardiac views in echocardiogram using histogram and statistical features},
\newblock \bibinfo{journal}{Procedia Computer Science} \bibinfo{volume}{46} (\bibinfo{year}{2015}) \bibinfo{pages}{1569--1576}.
\bibitem[{Pop(2020)}]{pop2020classification}
\bibinfo{author}{D.~Pop}, \bibinfo{title}{Classification of heart views in ultrasound images}, \bibinfo{year}{2020}.
\bibitem[{{\O}stvik et~al.(2019){\O}stvik, Smistad, Aase, Haugen, and Lovstakken}]{ostvik2019real}
\bibinfo{author}{A.~{\O}stvik}, \bibinfo{author}{E.~Smistad}, \bibinfo{author}{S.~A. Aase}, \bibinfo{author}{B.~O. Haugen}, \bibinfo{author}{L.~Lovstakken},
\newblock \bibinfo{title}{Real-time standard view classification in transthoracic echocardiography using convolutional neural networks},
\newblock \bibinfo{journal}{Ultrasound in medicine \& biology} \bibinfo{volume}{45} (\bibinfo{year}{2019}) \bibinfo{pages}{374--384}.
\bibitem[{Blaivas and Blaivas(2020)}]{blaivas2020all}
\bibinfo{author}{M.~Blaivas}, \bibinfo{author}{L.~Blaivas},
\newblock \bibinfo{title}{Are all deep learning architectures alike for point-of-care ultrasound?: evidence from a cardiac image classification model suggests otherwise},
\newblock \bibinfo{journal}{Journal of Ultrasound in Medicine} \bibinfo{volume}{39} (\bibinfo{year}{2020}) \bibinfo{pages}{1187--1194}.
\bibitem[{Gao et~al.(2021)Gao, Zhu, Liu, Hu, Yu, and Guo}]{gao2021automated}
\bibinfo{author}{Y.~Gao}, \bibinfo{author}{Y.~Zhu}, \bibinfo{author}{B.~Liu}, \bibinfo{author}{Y.~Hu}, \bibinfo{author}{G.~Yu}, \bibinfo{author}{Y.~Guo},
\newblock \bibinfo{title}{Automated recognition of ultrasound cardiac views based on deep learning with graph constraint},
\newblock \bibinfo{journal}{Diagnostics} \bibinfo{volume}{11} (\bibinfo{year}{2021}) \bibinfo{pages}{1177}.
\bibitem[{Deheyab et~al.(2022)Deheyab, Alwan, Rezzaqe, Mahmood, Hammadi, Kareem, and Ibrahim}]{deheyab2022overview}
\bibinfo{author}{A.~O.~A. Deheyab}, \bibinfo{author}{M.~H. Alwan}, \bibinfo{author}{I.~k.~A. Rezzaqe}, \bibinfo{author}{O.~A. Mahmood}, \bibinfo{author}{Y.~I. Hammadi}, \bibinfo{author}{A.~N. Kareem}, \bibinfo{author}{M.~Ibrahim},
\newblock \bibinfo{title}{An overview of challenges in medical image processing},
\newblock in: \bibinfo{booktitle}{Proceedings of the 6th International Conference on Future Networks \& Distributed Systems}, pp. \bibinfo{pages}{511--516}.
\bibitem[{Bharti and Mittal(2020)}]{bharti2020ultrasound}
\bibinfo{author}{P.~Bharti}, \bibinfo{author}{D.~Mittal},
\newblock \bibinfo{title}{An ultrasound image enhancement method using neutrosophic similarity score},
\newblock \bibinfo{journal}{Ultrasonic Imaging} \bibinfo{volume}{42} (\bibinfo{year}{2020}) \bibinfo{pages}{271--283}.
\bibitem[{Zhang et~al.(2021)Zhang, Wang, Jiang, Dong, Yi, and Hou}]{zhang2021cnn}
\bibinfo{author}{S.~Zhang}, \bibinfo{author}{Y.~Wang}, \bibinfo{author}{J.~Jiang}, \bibinfo{author}{J.~Dong}, \bibinfo{author}{W.~Yi}, \bibinfo{author}{W.~Hou},
\newblock \bibinfo{title}{Cnn-based medical ultrasound image quality assessment},
\newblock \bibinfo{journal}{Complexity} \bibinfo{volume}{2021} (\bibinfo{year}{2021}) \bibinfo{pages}{1--9}.
\bibitem[{Wu et~al.(2017)Wu, Cheng, Li, Lei, Wang, and Ni}]{wu2017fuiqa}
\bibinfo{author}{L.~Wu}, \bibinfo{author}{J.-Z. Cheng}, \bibinfo{author}{S.~Li}, \bibinfo{author}{B.~Lei}, \bibinfo{author}{T.~Wang}, \bibinfo{author}{D.~Ni},
\newblock \bibinfo{title}{Fuiqa: fetal ultrasound image quality assessment with deep convolutional networks},
\newblock \bibinfo{journal}{IEEE transactions on cybernetics} \bibinfo{volume}{47} (\bibinfo{year}{2017}) \bibinfo{pages}{1336--1349}.
\bibitem[{Adamson et~al.(2020)Adamson, Morris, Sun~Woan, Ma, Schnobrich, and Soni}]{adamson2020development}
\bibinfo{author}{R.~Adamson}, \bibinfo{author}{A.~E. Morris}, \bibinfo{author}{J.~Sun~Woan}, \bibinfo{author}{I.~W. Ma}, \bibinfo{author}{D.~Schnobrich}, \bibinfo{author}{N.~J. Soni},
\newblock \bibinfo{title}{Development of a focused cardiac ultrasound image acquisition assessment tool},
\newblock \bibinfo{journal}{ATS scholar} \bibinfo{volume}{1} (\bibinfo{year}{2020}) \bibinfo{pages}{260--277}.
\bibitem[{Madani et~al.(2018)Madani, Arnaout, Mofrad, and Arnaout}]{madani2018fast}
\bibinfo{author}{A.~Madani}, \bibinfo{author}{R.~Arnaout}, \bibinfo{author}{M.~Mofrad}, \bibinfo{author}{R.~Arnaout},
\newblock \bibinfo{title}{Fast and accurate view classification of echocardiograms using deep learning},
\newblock \bibinfo{journal}{NPJ digital medicine} \bibinfo{volume}{1} (\bibinfo{year}{2018}) \bibinfo{pages}{6}.
\bibitem[{Ebadi et~al.(2021)Ebadi, Xi, MacLean, Tremblay, Kohli, and Wong}]{DBLP:journals/corr/abs-2103-10003}
\bibinfo{author}{A.~Ebadi}, \bibinfo{author}{P.~Xi}, \bibinfo{author}{A.~MacLean}, \bibinfo{author}{S.~Tremblay}, \bibinfo{author}{S.~Kohli}, \bibinfo{author}{A.~Wong},
\newblock \bibinfo{title}{Covidx-us - an open-access benchmark dataset of ultrasound imaging data for ai-driven {COVID-19} analytics},
\newblock \bibinfo{journal}{CoRR} \bibinfo{volume}{abs/2103.10003} (\bibinfo{year}{2021}).
\bibitem[{Pedraza et~al.(2015)Pedraza, Vargas, Narv{\'a}ez, Dur{\'a}n, Mu{\~n}oz, and Romero}]{pedraza2015open}
\bibinfo{author}{L.~Pedraza}, \bibinfo{author}{C.~Vargas}, \bibinfo{author}{F.~Narv{\'a}ez}, \bibinfo{author}{O.~Dur{\'a}n}, \bibinfo{author}{E.~Mu{\~n}oz}, \bibinfo{author}{E.~Romero},
\newblock \bibinfo{title}{An open access thyroid ultrasound image database},
\newblock in: \bibinfo{booktitle}{10th International symposium on medical information processing and analysis}, volume \bibinfo{volume}{9287}, \bibinfo{organization}{SPIE}, pp. \bibinfo{pages}{188--193}.
\bibitem[{Singla et~al.(2023)Singla, Ringstrom, Hu, Lessoway, Reid, Nguan, and Rohling}]{DBLP:conf/miccai/SinglaRHLRNR23}
\bibinfo{author}{R.~Singla}, \bibinfo{author}{C.~Ringstrom}, \bibinfo{author}{G.~Hu}, \bibinfo{author}{V.~A. Lessoway}, \bibinfo{author}{J.~Reid}, \bibinfo{author}{C.~Y. Nguan}, \bibinfo{author}{R.~Rohling},
\newblock \bibinfo{title}{The open kidney ultrasound data set},
\newblock in: \bibinfo{editor}{B.~Kainz}, \bibinfo{editor}{J.~A. Noble}, \bibinfo{editor}{J.~A. Schnabel}, \bibinfo{editor}{B.~Khanal}, \bibinfo{editor}{J.~P. M{\"{u}}ller}, \bibinfo{editor}{T.~G. Day} (Eds.), \bibinfo{booktitle}{Simplifying Medical Ultrasound - 4th International Workshop, {ASMUS} 2023, Held in Conjunction with {MICCAI} 2023, Vancouver, BC, Canada, October 8, 2023, Proceedings}, volume \bibinfo{volume}{14337} of \textit{\bibinfo{series}{Lecture Notes in Computer Science}}, \bibinfo{publisher}{Springer}, \bibinfo{year}{2023}, pp. \bibinfo{pages}{155--164}.
\bibitem[{G{\'o}mez-Flores et~al.(2023)G{\'o}mez-Flores, Gregorio-Calas, and Coelho~de Albuquerque~Pereira}]{gomez2023bus}
\bibinfo{author}{W.~G{\'o}mez-Flores}, \bibinfo{author}{M.~J. Gregorio-Calas}, \bibinfo{author}{W.~Coelho~de Albuquerque~Pereira},
\newblock \bibinfo{title}{Bus-bra: A breast ultrasound dataset for assessing computer-aided diagnosis systems},
\newblock \bibinfo{journal}{Medical Physics}  (\bibinfo{year}{2023}).
\bibitem[{Leclerc et~al.(2019)Leclerc, Smistad, Pedrosa, {\O}stvik, Cervenansky, Espinosa, Espeland, Berg, Jodoin, Grenier, Lartizien, D'hooge, L{\o}vstakken, and Bernard}]{DBLP:journals/tmi/LeclercSPOCEEBJ19}
\bibinfo{author}{S.~Leclerc}, \bibinfo{author}{E.~Smistad}, \bibinfo{author}{J.~Pedrosa}, \bibinfo{author}{A.~{\O}stvik}, \bibinfo{author}{F.~Cervenansky}, \bibinfo{author}{F.~Espinosa}, \bibinfo{author}{T.~Espeland}, \bibinfo{author}{E.~A.~R. Berg}, \bibinfo{author}{P.~Jodoin}, \bibinfo{author}{T.~Grenier}, \bibinfo{author}{C.~Lartizien}, \bibinfo{author}{J.~D'hooge}, \bibinfo{author}{L.~L{\o}vstakken}, \bibinfo{author}{O.~Bernard},
\newblock \bibinfo{title}{Deep learning for segmentation using an open large-scale dataset in 2d echocardiography},
\newblock \bibinfo{journal}{{IEEE} Trans. Medical Imaging} \bibinfo{volume}{38} (\bibinfo{year}{2019}) \bibinfo{pages}{2198--2210}.
\bibitem[{Ouyang et~al.(2020)Ouyang, He, Ghorbani, Yuan, Ebinger, Langlotz, Heidenreich, Harrington, Liang, Ashley, and Zou}]{DBLP:journals/nature/OuyangHGYELHHLA20}
\bibinfo{author}{D.~Ouyang}, \bibinfo{author}{B.~He}, \bibinfo{author}{A.~Ghorbani}, \bibinfo{author}{N.~Yuan}, \bibinfo{author}{J.~Ebinger}, \bibinfo{author}{C.~P. Langlotz}, \bibinfo{author}{P.~A. Heidenreich}, \bibinfo{author}{R.~A. Harrington}, \bibinfo{author}{D.~H. Liang}, \bibinfo{author}{E.~A. Ashley}, \bibinfo{author}{J.~Y. Zou},
\newblock \bibinfo{title}{Video-based {AI} for beat-to-beat assessment of cardiac function},
\newblock \bibinfo{journal}{Nat.} \bibinfo{volume}{580} (\bibinfo{year}{2020}) \bibinfo{pages}{252--256}.
\bibitem[{Hemmsen et~al.(2010)Hemmsen, Petersen, Nikolov, Nielsen, and Jensen}]{hemmsen2010ultrasound}
\bibinfo{author}{M.~C. Hemmsen}, \bibinfo{author}{M.~M. Petersen}, \bibinfo{author}{S.~I. Nikolov}, \bibinfo{author}{M.~B. Nielsen}, \bibinfo{author}{J.~A. Jensen},
\newblock \bibinfo{title}{Ultrasound image quality assessment: A framework for evaluation of clinical image quality},
\newblock in: \bibinfo{booktitle}{Medical Imaging 2010: Ultrasonic Imaging, Tomography, and Therapy}, volume \bibinfo{volume}{7629}, \bibinfo{organization}{SPIE}, pp. \bibinfo{pages}{105--116}.
\bibitem[{Sfakianakis et~al.(2023)Sfakianakis, Simantiris, and Tziritas}]{DBLP:journals/bspc/SfakianakisST23}
\bibinfo{author}{C.~Sfakianakis}, \bibinfo{author}{G.~Simantiris}, \bibinfo{author}{G.~Tziritas},
\newblock \bibinfo{title}{{GUDU:} geometrically-constrained ultrasound data augmentation in u-net for echocardiography semantic segmentation},
\newblock \bibinfo{journal}{Biomed. Signal Process. Control.} \bibinfo{volume}{82} (\bibinfo{year}{2023}) \bibinfo{pages}{104557}.
\bibitem[{Liu et~al.(2021)Liu, Wang, Liu, Yang, and Tian}]{DBLP:journals/mia/LiuWLYT21}
\bibinfo{author}{F.~Liu}, \bibinfo{author}{K.~Wang}, \bibinfo{author}{D.~Liu}, \bibinfo{author}{X.~Yang}, \bibinfo{author}{J.~Tian},
\newblock \bibinfo{title}{Deep pyramid local attention neural network for cardiac structure segmentation in two-dimensional echocardiography},
\newblock \bibinfo{journal}{Medical Image Anal.} \bibinfo{volume}{67} (\bibinfo{year}{2021}) \bibinfo{pages}{101873}.
\bibitem[{Xu et~al.(2020)Xu, Liu, Shen, Wang, Liu, Wang, Wang, Li, Yu, Hou, Guo, Zhang, and He}]{DBLP:journals/cmig/XuLSWLWWLYHGZH20}
\bibinfo{author}{L.~Xu}, \bibinfo{author}{M.~Liu}, \bibinfo{author}{Z.~Shen}, \bibinfo{author}{H.~Wang}, \bibinfo{author}{X.~Liu}, \bibinfo{author}{X.~Wang}, \bibinfo{author}{S.~Wang}, \bibinfo{author}{T.~Li}, \bibinfo{author}{S.~Yu}, \bibinfo{author}{M.~Hou}, \bibinfo{author}{J.~Guo}, \bibinfo{author}{J.~Zhang}, \bibinfo{author}{Y.~He},
\newblock \bibinfo{title}{Dw-net: {A} cascaded convolutional neural network for apical four-chamber view segmentation in fetal echocardiography},
\newblock \bibinfo{journal}{Comput. Medical Imaging Graph.} \bibinfo{volume}{80} (\bibinfo{year}{2020}) \bibinfo{pages}{101690}.
\bibitem[{Barzegar et~al.(2021)Barzegar, Khatibi, and Hosseinsabet}]{barzegar2021proposing}
\bibinfo{author}{N.~Barzegar}, \bibinfo{author}{T.~Khatibi}, \bibinfo{author}{A.~Hosseinsabet},
\newblock \bibinfo{title}{Proposing novel methods for simultaneous cardiac cycle phase identification and estimating maximal and minimal left atrial volume (lav) from apical four-chamber view in 2-d echocardiography},
\newblock \bibinfo{journal}{Informatics in Medicine Unlocked} \bibinfo{volume}{23} (\bibinfo{year}{2021}) \bibinfo{pages}{100517}.
\bibitem[{Puig et~al.(2024)Puig, Friboulet, Ling, Varray, Mougharbel, Por{\'e}e, Provost, Garcia, and Millioz}]{puig2024boosting}
\bibinfo{author}{J.~Puig}, \bibinfo{author}{D.~Friboulet}, \bibinfo{author}{H.~J. Ling}, \bibinfo{author}{F.~Varray}, \bibinfo{author}{M.~Mougharbel}, \bibinfo{author}{J.~Por{\'e}e}, \bibinfo{author}{J.~Provost}, \bibinfo{author}{D.~Garcia}, \bibinfo{author}{F.~Millioz},
\newblock \bibinfo{title}{Boosting cardiac color doppler frame rates with deep learning},
\newblock \bibinfo{journal}{IEEE Transactions on Ultrasonics, Ferroelectrics, and Frequency Control}  (\bibinfo{year}{2024}).
\bibitem[{Lu et~al.(2023)Lu, Millioz, Varray, Por{\'e}e, Provost, Bernard, Garcia, and Friboulet}]{lu2023ultrafast}
\bibinfo{author}{J.~Lu}, \bibinfo{author}{F.~Millioz}, \bibinfo{author}{F.~Varray}, \bibinfo{author}{J.~Por{\'e}e}, \bibinfo{author}{J.~Provost}, \bibinfo{author}{O.~Bernard}, \bibinfo{author}{D.~Garcia}, \bibinfo{author}{D.~Friboulet},
\newblock \bibinfo{title}{Ultrafast cardiac imaging using deep learning for speckle-tracking echocardiography},
\newblock \bibinfo{journal}{IEEE Transactions on Ultrasonics, Ferroelectrics, and Frequency Control}  (\bibinfo{year}{2023}).
\bibitem[{Zamzmi et~al.(2022)Zamzmi, Rajaraman, Hsu, Sachdev, and Antani}]{zamzmi2022real}
\bibinfo{author}{G.~Zamzmi}, \bibinfo{author}{S.~Rajaraman}, \bibinfo{author}{L.-Y. Hsu}, \bibinfo{author}{V.~Sachdev}, \bibinfo{author}{S.~Antani},
\newblock \bibinfo{title}{Real-time echocardiography image analysis and quantification of cardiac indices},
\newblock \bibinfo{journal}{Medical image analysis} \bibinfo{volume}{80} (\bibinfo{year}{2022}) \bibinfo{pages}{102438}.
\bibitem[{Abdi et~al.(2017)Abdi, Luong, Tsang, Allan, Nouranian, Jue, Hawley, Fleming, Gin, Swift et~al.}]{abdi2017automatic}
\bibinfo{author}{A.~H. Abdi}, \bibinfo{author}{C.~Luong}, \bibinfo{author}{T.~Tsang}, \bibinfo{author}{G.~Allan}, \bibinfo{author}{S.~Nouranian}, \bibinfo{author}{J.~Jue}, \bibinfo{author}{D.~Hawley}, \bibinfo{author}{S.~Fleming}, \bibinfo{author}{K.~Gin}, \bibinfo{author}{J.~Swift}, et~al.,
\newblock \bibinfo{title}{Automatic quality assessment of apical four-chamber echocardiograms using deep convolutional neural networks},
\newblock in: \bibinfo{booktitle}{Medical Imaging 2017: Image Processing}, volume \bibinfo{volume}{10133}, \bibinfo{organization}{SPIE}, pp. \bibinfo{pages}{224--230}.
\bibitem[{Gaspari et~al.(2021)Gaspari, Teran, Kamilaris, and Gleeson}]{gaspari2021development}
\bibinfo{author}{R.~Gaspari}, \bibinfo{author}{F.~Teran}, \bibinfo{author}{A.~Kamilaris}, \bibinfo{author}{T.~Gleeson},
\newblock \bibinfo{title}{Development and validation of a novel image quality rating scale for echocardiography during cardiac arrest},
\newblock \bibinfo{journal}{Resuscitation Plus} \bibinfo{volume}{6} (\bibinfo{year}{2021}) \bibinfo{pages}{100097}.
\bibitem[{Sudharson and Kokil(2020)}]{sudharson2020ensemble}
\bibinfo{author}{S.~Sudharson}, \bibinfo{author}{P.~Kokil},
\newblock \bibinfo{title}{An ensemble of deep neural networks for kidney ultrasound image classification},
\newblock \bibinfo{journal}{Computer Methods and Programs in Biomedicine} \bibinfo{volume}{197} (\bibinfo{year}{2020}) \bibinfo{pages}{105709}.
\bibitem[{Wu et~al.(2018)Wu, Wang, and Wang}]{wu2018deep}
\bibinfo{author}{C.~Wu}, \bibinfo{author}{Y.~Wang}, \bibinfo{author}{F.~Wang},
\newblock \bibinfo{title}{Deep learning for ovarian tumor classification with ultrasound images},
\newblock in: \bibinfo{booktitle}{Advances in Multimedia Information Processing--PCM 2018: 19th Pacific-Rim Conference on Multimedia, Hefei, China, September 21-22, 2018, Proceedings, Part III 19}, \bibinfo{organization}{Springer}, pp. \bibinfo{pages}{395--406}.
\bibitem[{Zhu et~al.(2021)Zhu, Jin, Bao, Jiang, and Wang}]{zhu2021thyroid}
\bibinfo{author}{Y.-C. Zhu}, \bibinfo{author}{P.-F. Jin}, \bibinfo{author}{J.~Bao}, \bibinfo{author}{Q.~Jiang}, \bibinfo{author}{X.~Wang},
\newblock \bibinfo{title}{Thyroid ultrasound image classification using a convolutional neural network},
\newblock \bibinfo{journal}{Annals of translational medicine} \bibinfo{volume}{9} (\bibinfo{year}{2021}).
\bibitem[{He et~al.(2016)He, Zhang, Ren, and Sun}]{DBLP:conf/cvpr/HeZRS16}
\bibinfo{author}{K.~He}, \bibinfo{author}{X.~Zhang}, \bibinfo{author}{S.~Ren}, \bibinfo{author}{J.~Sun},
\newblock \bibinfo{title}{Deep residual learning for image recognition},
\newblock in: \bibinfo{booktitle}{2016 {IEEE} Conference on Computer Vision and Pattern Recognition, {CVPR} 2016, Las Vegas, NV, USA, June 27-30, 2016}, \bibinfo{publisher}{{IEEE} Computer Society}, \bibinfo{year}{2016}, pp. \bibinfo{pages}{770--778}.
\bibitem[{Zhang and Yang(2021)}]{zhang2021survey}
\bibinfo{author}{Y.~Zhang}, \bibinfo{author}{Q.~Yang},
\newblock \bibinfo{title}{A survey on multi-task learning},
\newblock \bibinfo{journal}{IEEE transactions on knowledge and data engineering} \bibinfo{volume}{34} (\bibinfo{year}{2021}) \bibinfo{pages}{5586--5609}.
\bibitem[{Simonyan and Zisserman(2014)}]{simonyan2014very}
\bibinfo{author}{K.~Simonyan}, \bibinfo{author}{A.~Zisserman},
\newblock \bibinfo{title}{Very deep convolutional networks for large-scale image recognition},
\newblock \bibinfo{journal}{arXiv preprint arXiv:1409.1556}  (\bibinfo{year}{2014}).
\bibitem[{Szegedy et~al.(2016)Szegedy, Vanhoucke, Ioffe, Shlens, and Wojna}]{szegedy2016rethinking}
\bibinfo{author}{C.~Szegedy}, \bibinfo{author}{V.~Vanhoucke}, \bibinfo{author}{S.~Ioffe}, \bibinfo{author}{J.~Shlens}, \bibinfo{author}{Z.~Wojna},
\newblock \bibinfo{title}{Rethinking the inception architecture for computer vision},
\newblock in: \bibinfo{booktitle}{Proceedings of the IEEE conference on computer vision and pattern recognition}, pp. \bibinfo{pages}{2818--2826}.
\bibitem[{Krizhevsky et~al.(2012)Krizhevsky, Sutskever, and Hinton}]{krizhevsky2012imagenet}
\bibinfo{author}{A.~Krizhevsky}, \bibinfo{author}{I.~Sutskever}, \bibinfo{author}{G.~E. Hinton},
\newblock \bibinfo{title}{Imagenet classification with deep convolutional neural networks},
\newblock \bibinfo{journal}{Advances in neural information processing systems} \bibinfo{volume}{25} (\bibinfo{year}{2012}).
\bibitem[{Suara et~al.(2023)Suara, Jha, Sinha, and Sekh}]{suara2023grad}
\bibinfo{author}{S.~Suara}, \bibinfo{author}{A.~Jha}, \bibinfo{author}{P.~Sinha}, \bibinfo{author}{A.~A. Sekh},
\newblock \bibinfo{title}{Is grad-cam explainable in medical images?},
\newblock in: \bibinfo{booktitle}{International Conference on Computer Vision and Image Processing}, \bibinfo{organization}{Springer}, pp. \bibinfo{pages}{124--135}.
\bibitem[{Selvaraju et~al.(2017)Selvaraju, Cogswell, Das, Vedantam, Parikh, and Batra}]{selvaraju2017grad}
\bibinfo{author}{R.~R. Selvaraju}, \bibinfo{author}{M.~Cogswell}, \bibinfo{author}{A.~Das}, \bibinfo{author}{R.~Vedantam}, \bibinfo{author}{D.~Parikh}, \bibinfo{author}{D.~Batra},
\newblock \bibinfo{title}{Grad-cam: Visual explanations from deep networks via gradient-based localization},
\newblock in: \bibinfo{booktitle}{Proceedings of the IEEE international conference on computer vision}, pp. \bibinfo{pages}{618--626}.
\bibitem[{Chattopadhay et~al.(2018)Chattopadhay, Sarkar, Howlader, and Balasubramanian}]{chattopadhay2018grad}
\bibinfo{author}{A.~Chattopadhay}, \bibinfo{author}{A.~Sarkar}, \bibinfo{author}{P.~Howlader}, \bibinfo{author}{V.~N. Balasubramanian},
\newblock \bibinfo{title}{Grad-cam++: Generalized gradient-based visual explanations for deep convolutional networks},
\newblock in: \bibinfo{booktitle}{2018 IEEE winter conference on applications of computer vision (WACV)}, \bibinfo{organization}{IEEE}, pp. \bibinfo{pages}{839--847}.

\end{thebibliography}

\end{document}